\documentclass[twoside]{article}

\usepackage[preprint]{aistats2026}
\usepackage{natbib} 
\usepackage{mathtools} 
\usepackage{booktabs} 
\usepackage{tikz} 
\usepackage{amsfonts,bm}
\usepackage{amsmath}
\usepackage{amssymb}
\usepackage{mathtools}
\usepackage{amsthm}
\usepackage[american]{babel}
\usepackage{multirow}
\usepackage{hyperref}
\usepackage{url}
\usepackage{graphicx}
\usepackage{algorithm, algorithmic}
\usepackage{rotating}
\usepackage{subcaption}
\usepackage{placeins}
\usepackage{subcaption}
\usepackage{afterpage} 
\usepackage{adjustbox}
 
\newcommand{\INDSTATE}[1][1]{\STATE\hspace{#1\algorithmicindent}}

\begin{document}

\twocolumn[

\aistatstitle{DeepRV: Accelerating Spatiotemporal Inference with Pre-trained Neural Priors}

\aistatsauthor{
  Jhonathan Navott$^{1,2*}$ \And
  Daniel Jenson$^{2,*}$ \And
  Seth Flaxman$^{2}$ \And
  Elizaveta Semenova$^{1, \dagger}$ 
}

\aistatsaddress{
  $^{1}$School of Public Health, Imperial College London, UK \\
  $^{2}$Department of Computer Science, University of Oxford, UK
} ]
\begingroup
\renewcommand\thefootnote{}\footnotetext{
${}^*$Equal contribution.\quad
${}^\dagger$Corresponding author: \texttt{elizaveta.p.semenova@gmail.com}.\quad
Accepted at AISTATS 2026.
}
\addtocounter{footnote}{-1}
\endgroup

\begin{abstract}
Gaussian Processes (GPs) provide a flexible and statistically principled foundation for modelling spatiotemporal phenomena, but their $\mathcal{O}(N^3)$ scaling makes them intractable for large datasets. Approximate methods such as variational inference (VI), inducing-point (sparse) GPs, low-rank kernel approximations (\textit{e.g.}, Nystr\"om methods and random Fourier features), and  approximations such as INLA improve scalability but typically trade off accuracy, calibration, or modelling flexibility. We introduce \textbf{DeepRV}, a neural-network surrogate that \textit{replaces GP prior sampling}, while closely matching full GP accuracy at inference including hyperparameter estimates, and reducing computational complexity to $\mathcal{O}(N^2)$, increasing scalability and inference speed. DeepRV serves as a drop-in replacement for GP prior realisations in \textit{e.g.}~MCMC-based probabilistic programming pipelines, preserving full model flexibility. Across simulated benchmarks, non-separable spatiotemporal GPs, and a real-world application to education deprivation in London (n = 4,994 locations), DeepRV achieves the highest fidelity to exact GPs while substantially accelerating inference. 
Code is provided in the \href{https://github.com/MLGlobalHealth/dl4bi}{\texttt{dl4bi}} Python package, with all experiments run on a single consumer-grade GPU to ensure accessibility for practitioners.

\end{abstract}

\begin{figure*}[t]
\centering
\includegraphics[width=\textwidth]{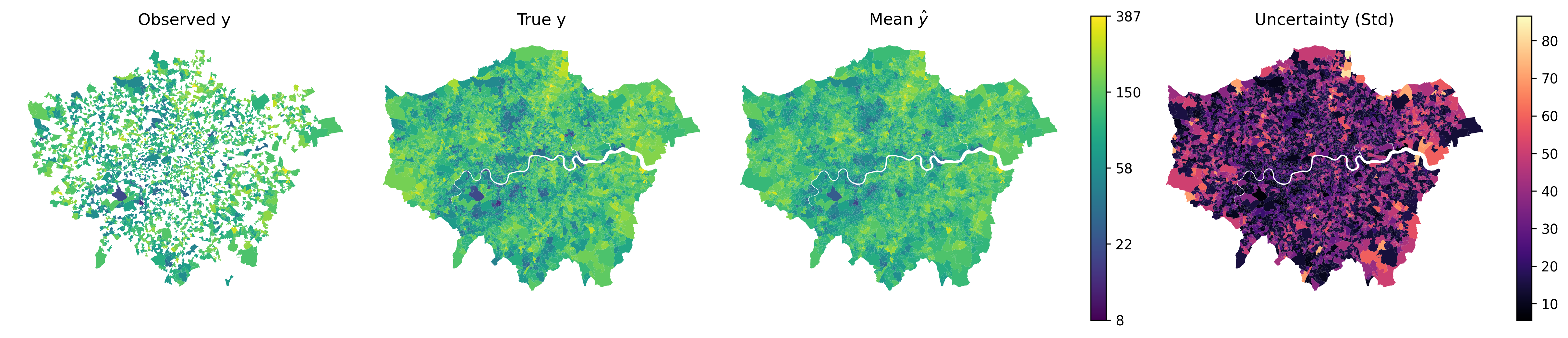}
\caption{DeepRV predictive evaluation on the London LSOA education deprivation dataset (= 4,994 locations). Panels show (from left to right): observed $\mathbf{y}$ (masked), full true $\mathbf{y}$, DeepRV posterior predictive mean $\hat{\mathbf{y}}$, and DeepRV posterior predictive uncertainty (standard deviation).}
\label{fig:lsoa_obs_means}
\end{figure*}

\section{Introduction}
GPs provide a principled Bayesian framework for modelling spatial and spatiotemporal phenomena, offering both predictive accuracy and uncertainty quantification. Their nonparametric nature allows GPs to flexibly capture complex nonlinear relationships without strong assumptions about functional form, while kernel design encodes spatial correlations and domain knowledge. These strengths have driven adoption in disease mapping \citep{Diggle1998Model,DiggleGiorgi2015PrevalenceMapping,ZhouJi2020TransmissionCOVID,SpatialLogGaussianCoxDiggle2013,Lawson2018Bayesian},
air pollution modelling 
\citep{Desai2022DeepGP_AQ,Patel2022ScalableGP_AQ,Wang2021GP_Uncertainty_AQ,Cheng2014FineGrained_GP_AQ,Stoddart2023GP_Kampala,Sonabend2024MixtureGP_AQHealth}, and climate risk analysis \citep{Mansour2024DMS_Mediterranean,Agou2022WarpedGP_Precip,Klockmann2024VarianceGP_Climate,Xiong2021GP_DataFusion,Wang2024FloodGP_Inundation,Koh2021Wildfire_LGCP}.
Importantly, GPs yield interpretable posteriors that enhance decision-making under uncertainty.

As datasets grow, the $\mathcal{O}(N^3)$ cost of GPs renders them computationally infeasible. Approximations such as inducing points \citep{Csato2002Sparse,Snelson2006Sparse,Quinonero2005Unifying,Titsias2009Variational}, low-rank factorizations \textit{e.g.}~random Fourier features (RFFs) \citep{Rahimi2007RFF}, variational inference (VI) \citep{Hensman2013Scalable,Hensman2015Scalable,Matthews2017SVGPConvergence}, and the Integrated Nested Laplace Approximation (INLA) \citep{Rue2009INLA,Rue2017INLAReview} enable more scalability, but each trades accuracy for efficiency or imposes restrictive modelling assumptions.

Neural surrogates such as PriorVAE \citep{semenova2022priorvae}, PriorCVAE \citep{semenova2023priorcvae}, and $\pi$VAE \citep{mishra2020pi} offer an alternative path, replacing the GP prior with a learned generative decoder to balance flexibility and scalability. These models avoid the cubic cost of exact GP inference, with complexity determined by the decoder architecture (typically quadratic in the number of locations), but often sacrifice accuracy.\\

\textbf{DeepRV} provides an alternative and elegant neural surrogate approach with very high fidelity to full GP inference while substantially improving scalability and speed. We summarise our contributions as follows:
\begin{enumerate}
\item The novel \textbf{DeepRV} training paradigm for emulating GPs.  
\item A comprehensive set of experiments benchmarking INLA, PriorCVAE, RFFs, ADVI, inducing points and DeepRV on 2D Gaussian processes. DeepRV achieves the highest fidelity to full GP Markov Chain Monte Carlo (MCMC) across predictive and parameter metrics, while accelerating MCMC inference substantially.
\item Applying DeepRV to non-separable spatiotemporal GPs, where it flexibly handles covariance structures challenging for INLA and RFFs.
\item Evaluating DeepRV on the education dimension of deprivation in London at the LSOA level (n = 4,994 locations), where standard GP approaches are computationally prohibitive.
\end{enumerate}

Below we review background and related work, introduce DeepRV, and evaluate it across a range of benchmarks and a real-life dataset.

\begin{figure*}[t]
\centering
\includegraphics[width=0.80\textwidth]{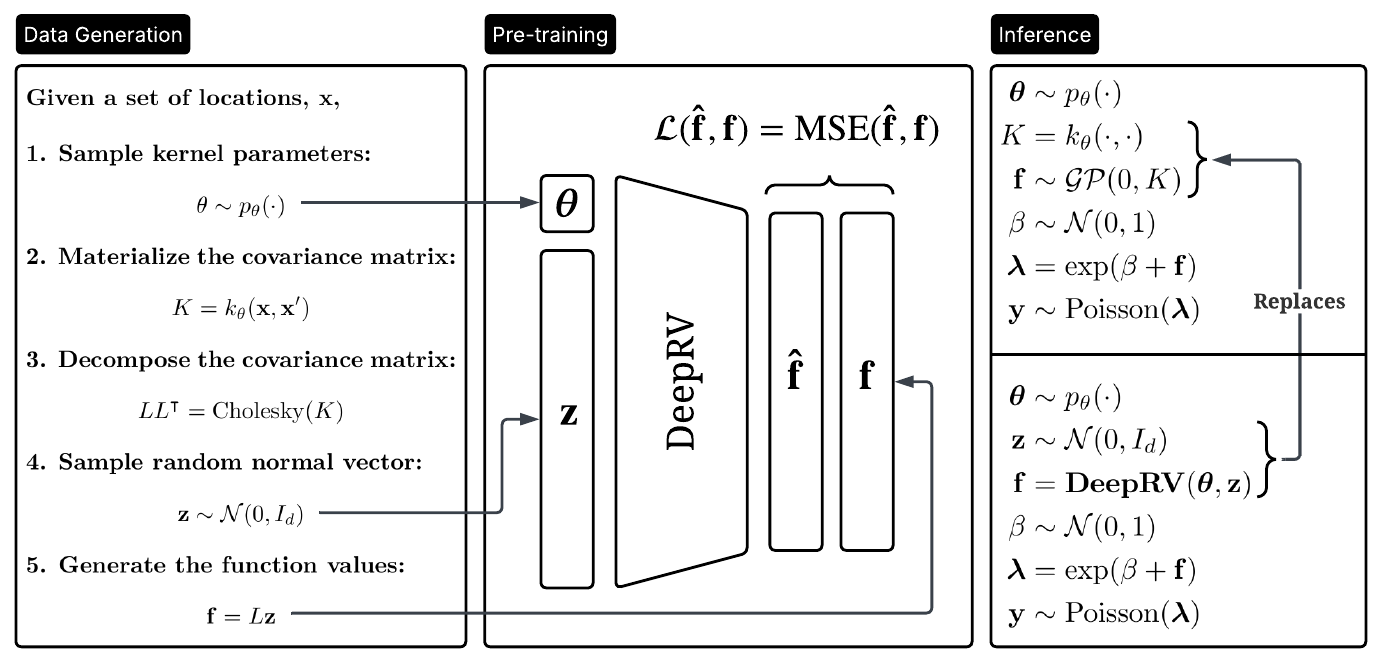}
\caption{A schematic overview of the DeepRV approach. Left panel details the data generating process used for pre-training. The middle panel shows the input and output of DeepRV during pre-training. In the right panel are two statistical models for inference, the first representing a traditional model that uses a GP prior, while the second one replaces the GP with a trained DeepRV network. In practice, for a given spatiotemporal setting, data generation and pre-training are a one-off cost which takes a few minutes to hours. Once complete, inference runtime is substantially sped-up, and can be repeated as many times as necessary.}
\label{fig:deep-rv-flow}
\end{figure*}

\section{Background}
\subsection{Gaussian Processes (GPs)}
A GP is an infinite collection of random variables, any finite subset of which has a joint multivariate Gaussian distribution \citep{williams2006gaussian}. Formally, a stochastic process $\{f(x): x\in\mathcal{X}\}$ is a GP if for any finite set of inputs, $x_1,\ldots, x_n\in\mathcal{X}$, the random vector
\begin{equation}
\mathbf{f}=(f(x_1),\ldots,f(x_n))^\intercal
\label{eq:gp1}
\end{equation}
is distributed as:
\begin{equation}
\mathbf{f}\sim\mathcal{N}(\mu(\mathbf{x}),K(\mathbf{x},\mathbf{x}^\prime))
\label{eq:gp2}
\end{equation}
where $\mu(\mathbf{x}) = [\mu(x_1),\ldots,\mu(x_n)]^\intercal$ is the mean function and $K(\mathbf{x},\mathbf{x}^\prime)$ is the covariance matrix with entries $K_{ij}=k_\theta(x_i,x_j)$, defined by a positive semidefinite kernel function $k_\theta:\mathcal{X}\times\mathcal{X}\to\mathbb{R}$ parametrised by $\theta$, which is often the tuple of lengthscale and variance $\theta=(\ell, \sigma^2)$. Thus, a GP can be written as
\begin{equation}
f(x)\sim\mathcal{GP}(\mu(x),k_\theta(x,x^\prime))
\label{eq:gp3}
\end{equation}
The kernel function $k_\theta(x, x^\prime)$ plays a central role in controlling the smoothness, periodicity, and other structural properties of the functions drawn from the GP prior. Commonly used kernels include the squared exponential / radial basis function, Mat\'ern family, periodic kernel, and linear / polynomial kernel. Combining these kernels through addition and multiplication allow practitioners to model highly structured signals.

\subsection{GPs for Spatiotemporal Inference}
GPs have become a central tool for spatial and spatiotemporal inference, providing a flexible probabilistic framework. In a Bayesian formulation, a GP prior models latent functions such as disease risk, pollution concentration, or meteorological variables over geographical and temporal domains.  By defining a covariance structure that encodes correlation, typically as a function of distance, GPs enable coherent interpolation from sparse and irregularly spaced observations to unobserved locations, a task often referred to as kriging  \citep{Diggle1998Model}. The posterior predictive distribution not only yields point estimates but also quantifies uncertainty, making GPs especially valuable for risk-sensitive applications. 
Their capacity to integrate prior knowledge through kernel design allows GPs to capture domain-specific structure, while approximate inference methods extend their applicability to increasingly large spatial and spatiotemporal datasets. In the subsequent section we review a variety of techniques used to scale GPs for spatial and spatiotemporal inference.

\section{Related Work}
In this section we detail the leading techniques used to scale GPs for inference tasks. We provide a qualitative comparison in \autoref{tab:method-comparison}. Further details on these methods, including their specific implementations used in this paper, are provided in \autoref{app:gp-approximations}.

\subsection{INLA} Integrated Nested Laplace Approximation (INLA) provides a deterministic alternative to MCMC for latent Gaussian models, whose computational cost can be prohibitive for high-dimensional structured settings \citep{Rue2009INLA}. INLA approximates posterior marginals via nested Laplace approximations coupled with deterministic numerical integration. By exploiting the sparse precision matrices of Gaussian Markov random fields (GMRFs), it enables scalable inference for hierarchical latent Gaussian models widely used in spatial statistics, disease mapping, and environmental risk assessment \citep{bakka2018spatial}. The stochastic partial differential equation (SPDE) formulation provides an explicit link between continuously indexed Gaussian fields and discrete GMRFs, facilitating large-scale spatial and spatiotemporal modelling \citep{Lindgren2011SPDE}. In practice, \texttt{R-INLA} \citep{Rue2017INLAReview} is the primary implementation, and the \texttt{inlabru} \citep{inlabru} package builds on it to support richer model specifications, including non-linear predictors via iterative linearisation. Despite these advantages, practical limitations remain: inference is primarily provided as marginal posterior summaries rather than full joint posteriors \citep{gomezrubio2017spatial}; the software supports a broad but finite catalogue of likelihoods and latent components, with additional families or extensions requiring non-trivial implementation effort \citep{inlamanual}; and genuinely non-separable space-time structures need specialised model formulations beyond default workflows \citep{bakka2018spatial}. The comparisons performed in this paper rely on the \texttt{R-INLA} interface.
\subsection{Sparse GPs}
Sparse GPs introduce a small set of inducing points, $M\ll N$, that are intended to summarize the full dataset, reducing complexity from $\mathcal{O}(N^3)$ to $\mathcal{O}(NM^2)$. Early formulations include pseudo-input GPs \citep{Snelson2006Sparse} and the unifying framework of \citet{Quinonero2005Unifying}, which approximate the covariance structure directly. \citet{Titsias2009Variational} provide a Bayesian framework for learning inducing variables and minimizing information loss, while \citet{Hensman2013Scalable} provide a stochastic variational inference extension that enables training on massive datasets using batch optimisation. These methods offer scalability, but often sacrifice accuracy. 

\begin{table}[h!]
\small\centering
\caption{Qualitative comparison of spatial inference techniques. Citations for each method are in the text.}
\label{tab:method-comparison}
\resizebox{\columnwidth}{!}{%
\begin{tabular}{lccc}
\toprule
\textbf{Method} & \textbf{Accuracy} & \textbf{Flexibility} & \textbf{Scalability} \\
\midrule
\textbf{GP} & High & High & Low \\
\textbf{INLA} & Med-High & Low–Med & High \\
\textbf{Inducing Points} & Med & Med & Med-High \\
\textbf{VI} & Med–Low & High & Med-High \\
\textbf{PriorCVAE} & Med & High & Med \\
\textbf{RFF} & Med-low & Med & Med-High \\
\textbf{DeepRV (Ours)} & High & High & Med \\
\bottomrule
\end{tabular}
} 
\end{table}
\subsection{Low-rank Factorizations}
A complementary approach to inducing points for scaling GPs is based on low-rank factorizations of the covariance matrix. The core idea is that many kernels have covariance matrices that are approximately low rank, particularly when the input data is smooth or lies on a low-dimensional data manifold. The Nystr\"om method \citep{Williams2001Nystrom} exploits a subset of columns of the kernel matrix to construct a low rank approximation. On the other hand, \cite{rahimi2007random} use random Fourier features (RFFs) to approximate shift-invariant kernels via Monte Carlo features drawn from the spectral density. These approaches reduce the $\mathcal{O}(N^3)$ to $\mathcal{O}(NM^2)$ or $\mathcal{O}(ND)$ where $M$ is the number of basis functions and $D$ is the number of random features. While highly scalable, these techniques are often less accurate than full GPs or INLA \citep{heaton2018casestudycompetitionmethods}.

\subsection{Variational Inference} Variational inference (VI) provides a scalable alternative to MCMC by approximating a computationally expensive posterior with a parametric distribution drawn from a restricted family. Rather than sampling from the exact posterior, VI fits this distribution by minimizing the KL divergence between the variational distribution and the true posterior. Commonly, the variational distribution is parametrized as a multivariate Gaussian \citep{Hensman2015Scalable,Matthews2017SVGPConvergence}. Modern probabilistic programming frameworks enable flexible VI methods, such as Automatic Differentiation Variational Inference (ADVI), which can be applied to GP models without model-specific derivations \citep{kucukelbir2017automatic,phan2019composable}. While VI scales well and integrates naturally with probabilistic models, its accuracy is inherently limited by the expressiveness of the variational family.
\subsection{Composite Likelihood Approximation}
Composite likelihood methods scale GPs by replacing the full joint likelihood with a product of low-dimensional conditional distributions. A prominent approach is the Vecchia approximation \citep{vecchia_1998}, which factorizes to:
\[
p(\mathbf{f}) \approx \prod_{i=1}^N p\big(f_i \mid \mathbf{f}_{C(i)}\big),
\]
where each conditioning set $C(i)$ contains only a small number of previously ordered locations. This induces a sparse, full-rank directed graphical structure and reduces computational complexity to $\mathcal{O}(Nm^2)$, with $m \ll N$. Subsequent work has shown that careful ordering and grouping strategies can substantially improve accuracy \citep{Guinness_2018,Katzfuss_2021}. However, despite being able to achieve great accuracy, these methods are inherently sequential which hinders scalability. 

\subsection{Neural Surrogates}
Recent literature proposes neural surrogates for GPs that can be used as a drop-in replacement in inference frameworks, and include PriorVAE, PriorCVAE, and $\pi$VAE \citep{semenova2022priorvae,semenova2023priorcvae,mishra2020pi}. All of these techniques share a common foundation in Variational autoencoders (VAEs). The fundamental idea of the VAE is that a collection of unknown latent variables control the target data generating process. When the prior on these latents is Gaussian, this is also known as a deep latent Gaussian model (DLGM) \citep{pml2book}.

The objective of these neural surrogates is to train a VAE that can generate samples from a Gaussian process prior. PriorVAE and PriorCVAE, the conditional variant, use a standard MLP-based encoder and decoder. Once the model has been trained, the decoder can generate samples from the prior by decoding a random latent vector, $\mathbf{z}$, and optional conditioning variables, such as the lengthscale and variance. 

These techniques, while fast and flexible, suffer from poor accuracy, largely due to the weaknesses associated with VAE-based architectures, such as posterior collapse and oversmoothing. Posterior collapse occurs during encoding, when the model ignores parts of the latent space and produces uninformative representations. Oversmoothing occurs during decoding, when the generated samples are excessively smooth.
Furthermore, these errors compound: an error in approximating the latent distribution is exacerbated by a lossy decoding process. These limitations motivated the design of DeepRV, which we detail next.
\begin{figure*}[t]
    \centering
    \small
    \begin{subfigure}[b]{0.49\textwidth}
        \includegraphics[width=\textwidth]{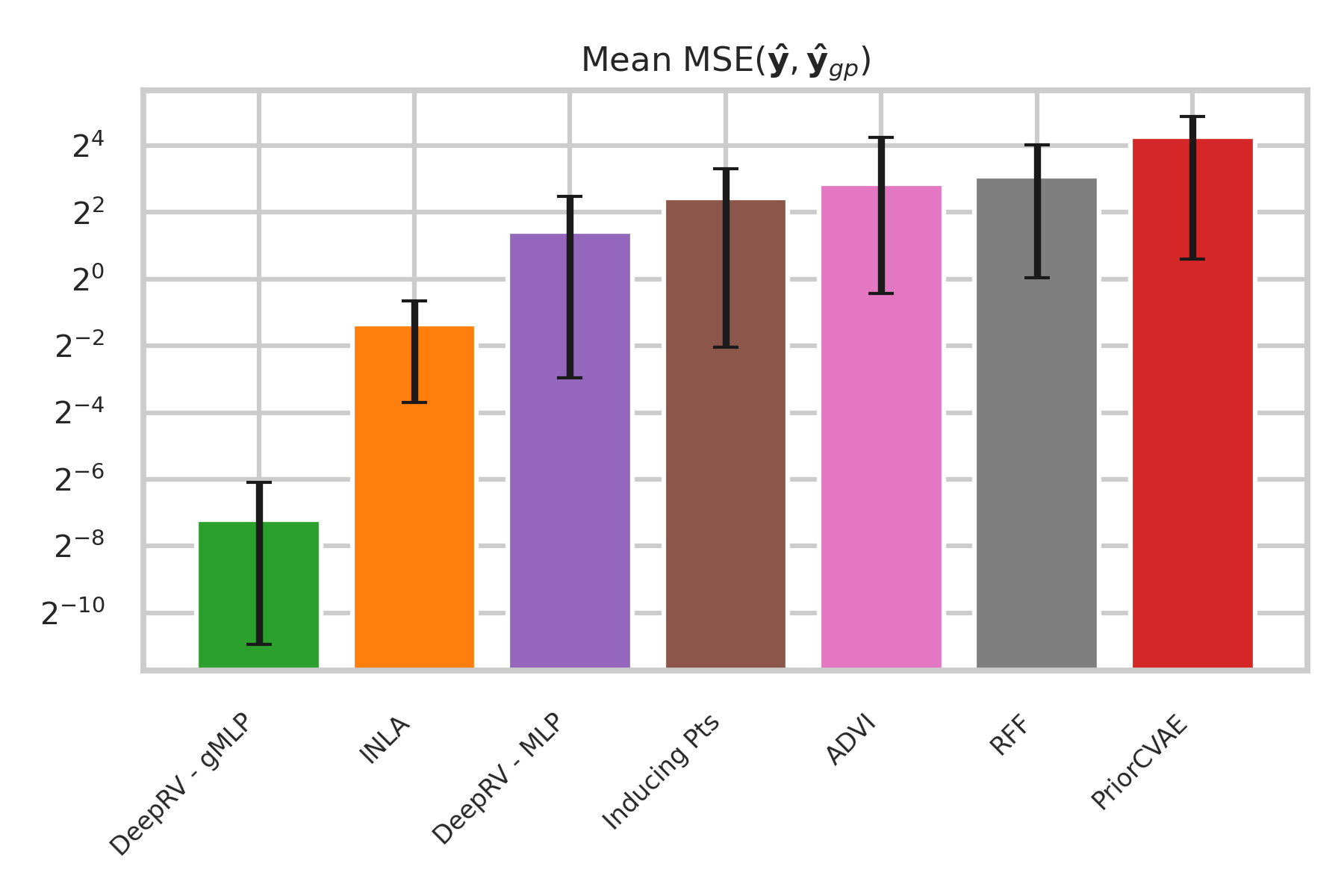}
    \end{subfigure}
    \begin{subfigure}[b]{0.49\textwidth}
        \includegraphics[width=\textwidth]{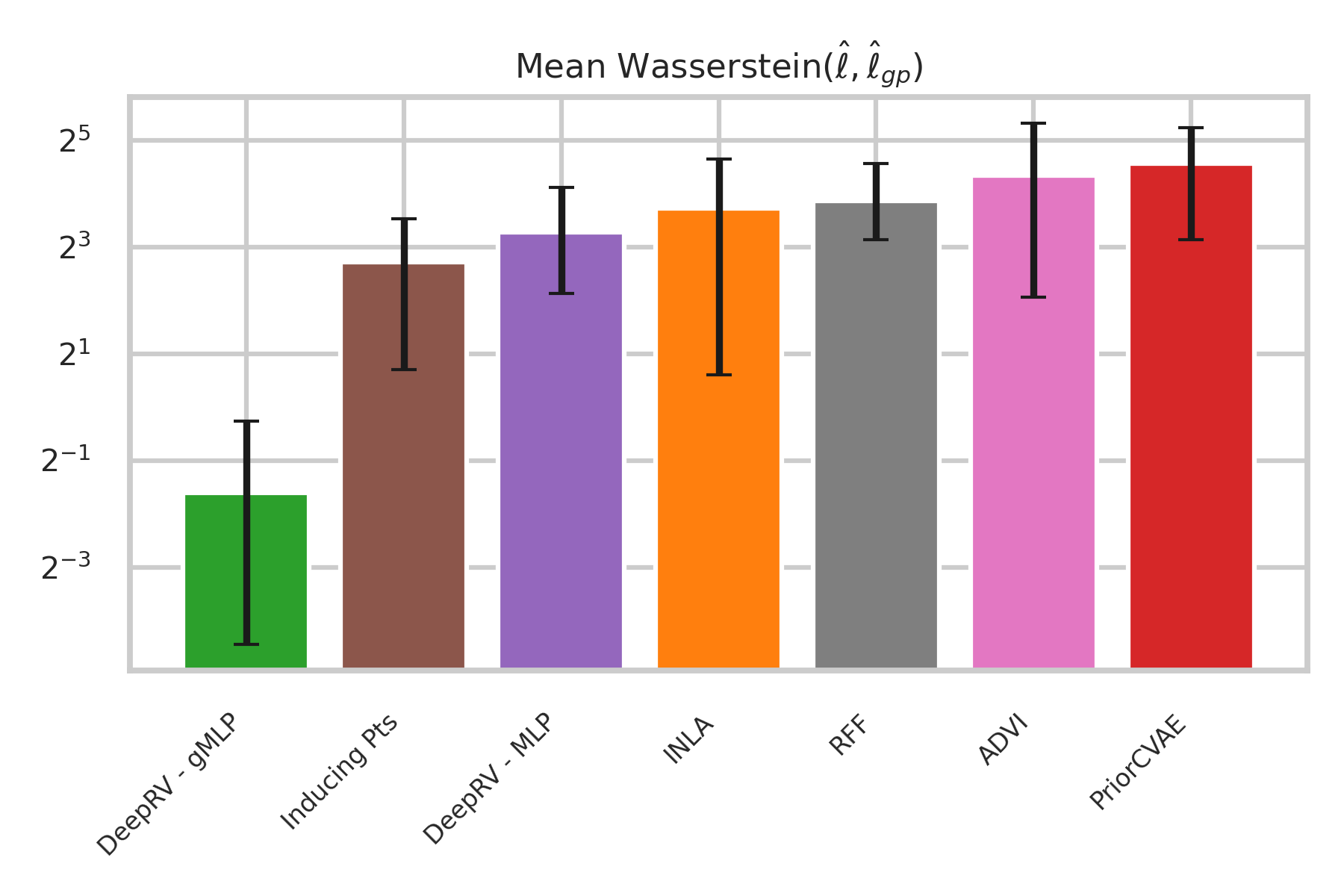}
    \end{subfigure}
    \caption{Mat\'ern-1/2 benchmarking results: (a) MSE of each model's posterior mean $\boldsymbol{\hat y}$ relative to the full GP MCMC posterior mean $\bm{\hat y}_\text{gp}$; (b) Wasserstein distance between each model's posterior over lengthscale $\hat \ell$ and full GP MCMC posterior over lengthscale $\hat\ell_{\mbox{gp}}$. Y axis is $\operatorname{log}_2$-scaled, to allow a clear comparison of all methods. Results are averaged across true lengthscales and grid sizes over 15 runs, with 10\% and 90\% quantiles reported.}
    \label{fig:bar_summary_main_matern_1_2}
\end{figure*}
\section{DeepRV}
\subsection{Method}
We introduce \textbf{DeepRV}, a highly accurate neural surrogate for Gaussian process realizations. DeepRV differs from previously described neural surrogates in three principal ways: (1) it eliminates the encoding process entirely, (2) it has no information bottleneck, and (3) it leverages factorized stochastic processes directly for training. 

These changes hinge on a key insight: for any stochastic process that decomposes into a latent random vector and a linear transformation, the encoding step can be entirely avoided, removing a source of error and allowing the model to focus on accurate decoding. This structure holds naturally for Gaussian processes, since realizations $f(x)\sim\mathcal{GP}(\mu(x),k_\theta(x,x^\prime))$ as in Equations~\eqref{eq:gp1}-\eqref{eq:gp3}
can be sampled as follows:
\begin{equation}\label{eqn:samplegp}
\begin{split}
\mathbf{z}&\sim \mathcal{N}(0, \mathbf{I}) \\
L&=\textbf{Cholesky}(K) \\
\mathbf{f}&:= \boldsymbol{\mu} + L\mathbf{z} \\
\end{split}
\end{equation}
Thus, with a known $\mathbf{z}$ and $\mathbf{f}$, we can train a network to learn $L$ and sample from the GP directly. This process is depicted visually in \autoref{fig:deep-rv-flow}. 
\subsection{Architectures}
In the following, we present 3 architecture variants for DeepRV: (1) a vanilla MLP, (2) a gated MLP (gMLP) \citep{liu2021pay}, and (3) a transformer \citep{vaswani2017attention}.
\subsubsection{MLP}
A multilayer perceptron (MLP) consists of sequential layers performing linear transformations followed by nonlinear activations. Stacking multiple layers enables the network to approximate complex, nonlinear mappings. For DeepRV, we use a simple two-layer MLP without dimensionality reduction and with ReLU activations. This maintains consistency with PriorCVAE and highlights that the performance gain achieved from the novel training procedure and decoder-only design, and not only the architectural complexity.

\begin{table*}[t]
\scriptsize
\centering
\setlength{\tabcolsep}{1.5pt}
\begin{tabular}{ll|ccccccc}
\toprule
Metric & N & GP & INLA & Inducing Pts & RFF & PriorCVAE & DeepRV-MLP & DeepRV-gMLP \\
\midrule
 & 256 & - & 0.392 ± 0.20 & 4.802 ± 2.87 & 7.363 ± 2.98 & 8.064 ± 4.37 & 1.116 ± 0.53 & \textbf{0.002 ± 0.00} \\
 & 576 & - & 0.333 ± 0.04 & 5.011 ± 1.93 & 9.790 ± 3.87 & 13.470 ± 8.89 & 2.373 ± 0.66 & \textbf{0.005 ± 0.00} \\
MSE($\mathbf{\hat{y}}_{gp}, \mathbf{\hat{y}}$) & 1024 & - & 0.411 ± 0.01 & 6.436 ± 1.65 & 9.735 ± 2.40 & 17.752 ± 4.42 & 4.895 ± 1.46 & \textbf{0.009 ± 0.00} \\
 & 2,304 & - & 0.261 ± 0.11 & 8.678 ± 5.97 & 11.296 ± 5.33 & 9.877 ± 4.39 & 3.714 ± 1.82 & \textbf{0.013 ± 0.01} \\
 & 4,096 & - & 0.570 ± 0.49 & 2.158 ± 1.83 & 4.083 ± 3.38 & 47.949 ± 46.96 & 1.367 ± 1.13 & \textbf{0.005 ± 0.00} \\
\midrule
 & 256 & - & 11.68 ± 4.29 & 4.87 ± 1.57 & 11.72 ± 4.49 & 15.66 ± 9.79 & 9.23 ± 2.74 & \textbf{0.13 ± 0.08} \\
 & 576 & - & 11.74 ± 5.08 & 5.40 ± 2.31 & 13.81 ± 4.57 & 25.62 ± 7.55 & 9.36 ± 1.53 & \textbf{0.21 ± 0.06} \\
Wass($\hat{\ell}_{gp}, \hat{\ell}$) & 1024 & - & 13.59 ± 7.94 & 9.95 ± 5.16 & 15.72 ± 5.83 & 26.64 ± 10.05 & 12.91 ± 6.29 & \textbf{0.26 ± 0.08} \\
 & 2,304 & - & 12.90 ± 5.92 & 7.16 ± 2.87 & 15.09 ± 3.65 & 27.33 ± 9.15 & 12.20 ± 3.71 & \textbf{0.44 ± 0.36} \\
 & 4,096 & - & 16.19 ± 7.37 & 5.61 ± 2.18 & 16.33 ± 4.08 & 23.40 ± 7.03 & 4.87 ± 1.60 & \textbf{0.61 ± 0.48} \\
\midrule
 & 256 & 14.38 ± 4.23 & - & \textbf{27.76 ± 8.15} & 0.00 ± 0.00 & \textbf{56.14 ± 30.51} & \textbf{37.32 ± 10.11} & 21.30 ± 6.13 \\
 & 576 & 3.19 ± 0.61 & - & 11.70 ± 5.65 & 0.00 ± 0.00 & \textbf{13.97 ± 3.31} & \textbf{14.34 ± 1.71} & 8.14 ± 1.15 \\
$\text{ESS} (\ell)/{\text{sec}}$ & 1024 & 1.33 ± 0.49 & - & \textbf{8.10 ± 4.68} & 0.00 ± 0.00 & \textbf{11.98 ± 5.13} & 0.87 ± 0.36 & 6.47 ± 2.44 \\
 & 2,304 & 0.35 ± 0.06 & - & 3.97 ± 1.54 & 0.02 ± 0.01 & \textbf{8.20 ± 2.85} & 2.32 ± 0.32 & 3.26 ± 0.67 \\
 & 4,096 & 0.13 ± 0.03 & - & \textbf{2.82 ± 0.94} & 0.01 ± 0.01 & \textbf{2.99 ± 1.46} & 0.84 ± 0.12 & \textbf{2.74 ± 0.60} \\
\midrule
 & 256  & 274 ± 79.89 & \textbf{2 ± 0.08} & 154 ± 31.56 & 1,314 ± 127.31 & 81 ± 15.86 & 98 ± 21.46 & 157 ± 51.35 \\
 & 576  & 949 ± 257.32 & \textbf{4 ± 0.06} & 316 ± 64.61 & 373 ± 36.30 & 171 ± 8.58 & 188 ± 8.05 & 332 ± 95.52 \\
Infer Time (s) & 1024 & 2,546 ± 805.77 & \textbf{7 ± 0.07} & 566 ± 184.98 & 1,028 ± 175.29 & 231 ± 13.62 & 309 ± 52.97 & 510 ± 177.81 \\
 & 2,304 & 7,476 ± 1,428.08 & \textbf{38 ± 1.53} & 862 ± 195.73 & 1,653 ± 501.73 & 334 ± 60.64 & 394 ± 23.93 & 778 ± 242.11 \\
 & 4,096 & 20,659 ± 3,887.21 & \textbf{95 ± 1.88} & 955 ± 177.34 & 3,848 ± 1,242.73 & 595 ± 111.21 & 974 ± 170.98 & 939 ± 169.77 \\
\bottomrule
\end{tabular}
\caption{Mat\'ern-1/2 benchmarking results: (a) Posterior predictive MSE relative to full GP MCMC; (b) Wasserstein distance between inferred and full GP MCMC lengthscale posteriors; (c) Effective $\ell$ sample size (ESS) per second; (d) Inference time in seconds. Results are shown for each grid size and are averaged across the three true lengthscales (10, 30, 50) over 15 runs, with the standard error reported.}
\label{tab:res_matern_1_2}
\end{table*}

\subsubsection{gMLP}\label{subsubsec:deepRV_gMLP}
Gated multilayer perceptrons extend standard MLPs by introducing a gating mechanism \citep{liu2021pay}. If $X\in\mathbb{R}^{N\times D}$ where $N$ is the number of observations or locations and $D$ is an embedding dimension, each gMLP block can be represented by the following equations:
\begin{equation}
Z=\sigma(XU),\quad\tilde{Z}=\text{spatial-gate}(Z),\quad Y=\tilde{Z}V \\
\end{equation}
where $U$ and $V$ are trainable linear projections and $\sigma$ is a nonlinearity such as a GELU. The gating function splits $Z$ into two along the channel dimension, yielding $Z_1$ and $Z_2$. $Z_2$ is then projected with learnable $W$ and $b$ and gated by $Z_1$, i.e.
\begin{equation}
Z_1, Z_2 = \text{split}(Z),\quad\tilde{Z}= Z_1\odot (WZ_2 + b)
\end{equation}
Gated MLPs are similar to transformer blocks in that they intersperse an attention-like mechanism, i.e. spatial gating, with a feedfoward network. The benefit of this architecture is that it can leverage highly optimized general matrix multiplication (GEMM) operations on the GPU, making it extremely fast to train. A downside, however, is that the number of tokens or locations is fixed.
For DeepRV, we use a simple two-layer gMLP without an information bottleneck.

\subsubsection{Transformer}\label{subsec:trans}
To handle variably sized and arbitrarily chosen spatial locations (Section~\ref{subsec: multi location}), we employ a transformer-based DeepRV decoder. Transformers, originally introduced for sequence modelling \citep{vaswani2017attention}, consist of stacked multi-headed attention and feedforward layers with residual connections. We use a similar approach with two main modifications. First, we include ID positional embeddings to encode an ordering over locations. Concretely, the input token at index $i$ is given by
\begin{equation}
\mathbf{x}_i \;=\; \mathbf{z}_i \;\Vert\; \mathrm{embed}_s(\mathbf{s}_i) \;\Vert\; \mathrm{embed}_{\mathrm{id}}(i),
\end{equation}
where $\mathbf{z}_i$ is the latent input, $\mathbf{s}_i$ denotes the spatial location, $\mathrm{embed}_s(\cdot)$ is a spatial embedding (\textit{e.g.}, identity or RFF positional encodings), and $\mathrm{embed}_{\mathrm{id}}(i)$ is a learned ID embedding associated with the input index. Without ID embeddings, the model struggled to learn with arbitrary input locations. We hypothesize that these embeddings help the transformer implicitly represent the Cholesky factor $L$, whose structure depends on the ordering of inputs. Second, we incorporate a kernel-based attention bias \citep{bsatnp}. The biased attention is defined as
\begin{equation}
\mathcal{K}(\mathbf{Q}, \mathbf{K})\mathbf{V}
\;\coloneqq\;
\text{softmax}\!\left(
\frac{\mathbf{Q}\mathbf{K}^\intercal}{\sqrt{d_k}}
\;+\;
\alpha\,\mathbf{K}_{\boldsymbol{\theta}}
\right)\mathbf{V},
\end{equation}
where $\mathbf{K}_{\boldsymbol{\theta}}$ is the GP kernel matrix conditioned on hyperparameters $\boldsymbol{\theta}$, and $\alpha$ is a learnable scalar per attention head. This bias injects prior GP structure into attention weights and improves the fidelity with which the network reconstructs GP realisations.

\section{Data Generation, Pre-training, and Inference}
In order to train DeepRV, a dataset consisting of tuples of $(\boldsymbol{\theta},\mathbf{z},\mathbf{f})$ is created according to the following process:
\begin{enumerate}
\item Sample kernel parameters: $\boldsymbol{\theta}\sim p_\theta(\cdot).$
\item Materialize the kernel: $K=k_\theta(\mathbf{x},\mathbf{x}^\prime).$
\item Decompose the kernel: $L=\text{Cholesky}(K). $
\item Sample random normal vector: $\mathbf{z}\sim \mathcal{N}(0, I_d). $
\item Generate the function values: $ \mathbf{f}=L\mathbf{z}. $
\end{enumerate}
The input to DeepRV is $(\boldsymbol{\theta},\mathbf{z})$ and it outputs an estimate of function values $\mathbf{\hat{f}}$. The loss function is MSE between $\mathbf{\hat{f}}$ and the true $\mathbf{f}$.

Once trained, DeepRV can map a latent random vector, $\mathbf{z}$, and kernel parameters, $\boldsymbol{\theta}$, to an instance of the target stochastic process conditioned on those parameters. Accordingly, inside a probabilistic programming language like NumPyro, sampling from a GP can be replaced with sampling a random normal vector, $\mathbf{z}$, and passing $(\boldsymbol{\theta},\mathbf{z})$ through DeepRV in order to generate the sample $\mathbf{\hat{f}}$. This process is detailed in \autoref{fig:deep-rv-flow}.

\begin{figure*}[t]
    \centering
    \small
    \includegraphics[width=\textwidth]{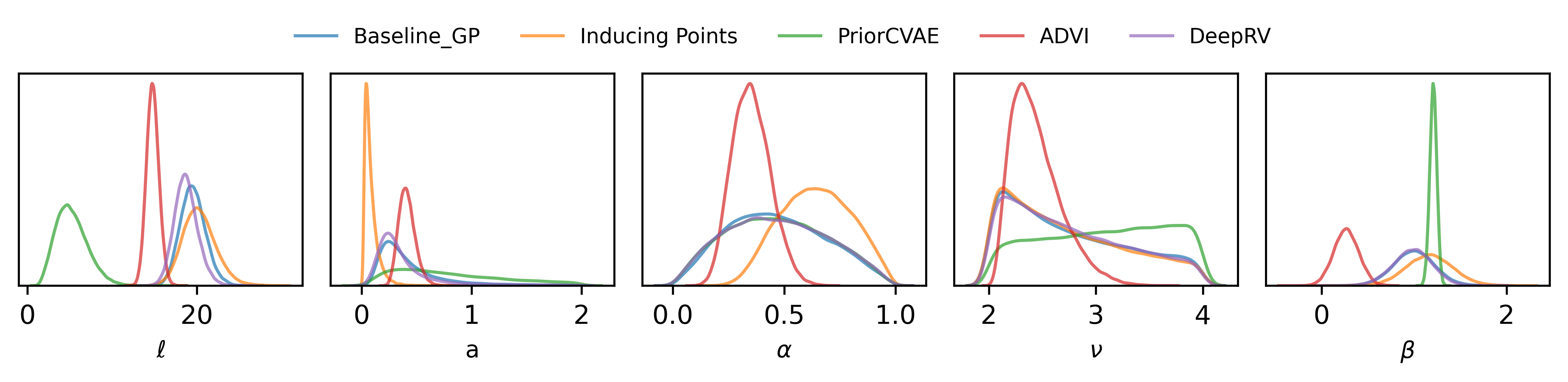}
    \caption{Spatiotemporal GP inferred hyperparameter posterior distributions. DeepRV closely matches GP on all hyperparameter posterior distributions.}
\label{fig:spatiotemp_kde}
\end{figure*}

\section{Experiments}\label{sec:exp}
\label{sec:experiments}
In this section, we present a set of experiments that evaluate DeepRV’s fidelity to full GP inference and its scalability by benchmarking against a range of existing methods. We further demonstrate its flexibility by performing inference with a non-separable spatiotemporal kernel, evaluate its applicability to real-world data, and introduce an extended variant capable of operating on arbitrary input locations.

All code required to reproduce these experiments is available in \href{https://github.com/MLGlobalHealth/dl4bi}{\texttt{dl4bi}} Python package.

\subsection{Benchmarking DeepRV}\label{subsec:benchmarking}
We simulated data over 2D grids of increasing resolution, $N=16^2$, $24^2$, $32^2$, $48^2$, and $64^2$, to assess the scalability and accuracy of DeepRV in spatial inference. We benchmarked DeepRV against INLA, Inducing Points, PriorCVAE, RFFs, and ADVI. INLA was tested using the standard \texttt{R-INLA} package, with meshes scaled to resolution, using the Laplace approximation with grid-based integration for accuracy. We selected $N^{2/3}$ anchors for the Inducing points method, and $2L$ features for RFF to match DeepRV’s complexity for fairness. For ADVI, we used NumPyro’s \texttt{AutoMultivariateNormal} guide to implement a full-rank Gaussian posterior. We refer the reader to Appendix \ref{app:gp-approximations} for implementation details of the benchmarking methods, and Appendix~\ref{subsec:appendix_vecchia} for a direct comparison between Vecchia approximations and DeepRV which were perform separately.

Grid coordinates were normalized to $[0,100]$ and used as GP inputs. DeepRV and PriorCVAE were trained to emulate a Matérn-1/2 GP prior, with mini-batches of 32 for 200K steps (300K for $48^2$ and $64^2$ grids). The lengthscale $\ell$ was drawn from a $\text{LogNormal}(3.0,0.4)$ prior (consistent with R-INLA mesh settings), and the variance fixed at 1, since the data can always be standardized prior to inference. After training, the learned priors were used in a NumPyro inference model with a Poisson likelihood:
\begin{align}\label{equation:Poissoin_GP_Eq}
{\boldsymbol{\theta}} :=& \ell, \sigma \sim p_\mathcal{\boldsymbol{\theta}}(\cdot), \notag\\
\mathbf{f}_{\boldsymbol{\theta}} \sim&\mathcal{GP}_{\boldsymbol{\theta}}(\cdot),\\
\beta \sim& \mathcal{N}(0, 1000),\notag\\
\boldsymbol{\lambda} =& \exp(\beta + \mathbf{f}_{\boldsymbol{\theta}}),\notag\\
\mathbf{y} \sim& \operatorname{Poisson}(\boldsymbol{\lambda}).\notag
\end{align}

For inference we used NUTS \citep{hoffman2014no} with two chains for grids $\leq 32^2$ and one chain for larger grids. While running a single chain is unusual, for the GP baseline at the largest grids one chain can take tens of hours, so this trade-off was necessary to make benchmarking feasible. We ran 4{,}000 warmup steps and 6{,}000 posterior draws per chain. Observations were generated with true $\beta=1.5$ and $\ell \in \{10,30,50\}$, with approximately 50\% masked in contiguous regions to increase difficulty. 

Results for the Matérn-1/2 kernel are presented in Table \ref{tab:res_matern_1_2} and Figure \ref{fig:bar_summary_main_matern_1_2}. We also repeated the experiment with a Matérn-3/2 kernel, which is not supported by standard \texttt{R-INLA} package for 2D inputs. Results are provided in Appendix Table \ref{tab:res_matern_3_2} and Figure \ref{fig:bar_summary_main_matern_3_2}.

Across settings, DeepRV achieves the highest fidelity to full GP inference in both predictive performance and hyperparameter recovery. INLA is consistently the fastest and provides competitive predictive accuracy, but weaker parameter inference. PriorCVAE yields the highest effective sample size per second, yet this is misleading since its predictive and parameter accuracy are among the lowest, highlighting ESS/sec as an incomplete standalone measure.

\subsection{DeepRV flexibility: non-separable spatiotemporal kernel}
To demonstrate DeepRV's flexibility relative to other GP approximation methods, we performed inference using a non-separable space–time covariance function inspired by Gneiting \citep{gneiting2002nonseparable}, defined as
\[
k_{\boldsymbol{\theta}}(\mathbf{s}, t; \mathbf{s}', t') = \frac{\sigma^2}{(a d_t^{2\alpha}+1)^{d/2}} \exp \Bigg(- \frac{\|\mathbf{s}-\mathbf{s}'\|^2}{\ell^2 (a d_t^{2\alpha}+1)^b} \Bigg)
\]
where $d_t = |t-t'|$, and the hyperparameters are ${\boldsymbol{\theta}} := \{\ell, \sigma^2, a, \alpha, b, \nu\}$. This kernel captures both spatial and temporal correlations in a non-separable manner.

Such genuinely non-separable structures are typically not directly supported in default INLA workflows and require specialised model formulations \citep{bakka2018spatial}, and they cannot be handled by standard RFF approximations. 

We followed the training and inference procedure described in Section~\ref{subsec:benchmarking}, with the only changes being the hyperparameter set $\boldsymbol{\theta}$ above, a single spatial grid of size $16^2$ with $5$ time steps, and we trained the neural networks for 500,000 steps. We set the hyperparameters to 
$
\sigma^2 = 1.0, \ell = 20.0, \beta = 1.0, a = 0.5, \alpha = 0.8, b = 1.0, \nu = 1.0.
$

Spatial masking was applied as before, with $\approx$50\% of observations masked in contiguous regions, consistent across all time steps. Additionally, observations at $t=2,3$ were removed to simulate partially observed temporal dynamics. The resulting inferred hyperparameter distributions are shown in Figure \ref{fig:spatiotemp_kde}, and the posterior predictive across time is presented in Figure \ref{fig:obs_means_spatio_temporal}. The results demonstrate that DeepRV matches GP predictive performance and parameter inference even in settings with more hyperparameters and complex interdependencies. This flexibility arises from DeepRV’s simple design, which does not rely on structural assumptions about the GP it emulates.

\subsection{Real-world application: London LSOA}
\label{subsec:london_lsoa}

We applied DeepRV to the education dimension of deprivation in London across 4,994 LSOAs. A household is deprived if no member has at least level 2 education and no one aged 16–18 is a full-time student. Data was taken from the ONS dataset generator\footnote{\url{https://www.ons.gov.uk/filters/dcf91941-1de3-4e26-8cae-4adec2a42f9c/dimensions}}, with boundaries from the ONS Open Geography Portal\footnote{\url{https://geoportal.statistics.gov.uk/}}

For validation, we also fit (i) a full GP at the MSOA level ($n=1{,}024$), where exact inference is still feasible, and (ii) a short full-GP run at the LSOA level (2 chains, 1{,}000 warmup, 500 posterior samples) to calibrate against DeepRV. This lets us check that DeepRV at both resolutions is consistent with a GP baseline.

\begin{figure}[h!]
\centering
\includegraphics[width=0.35\textwidth]{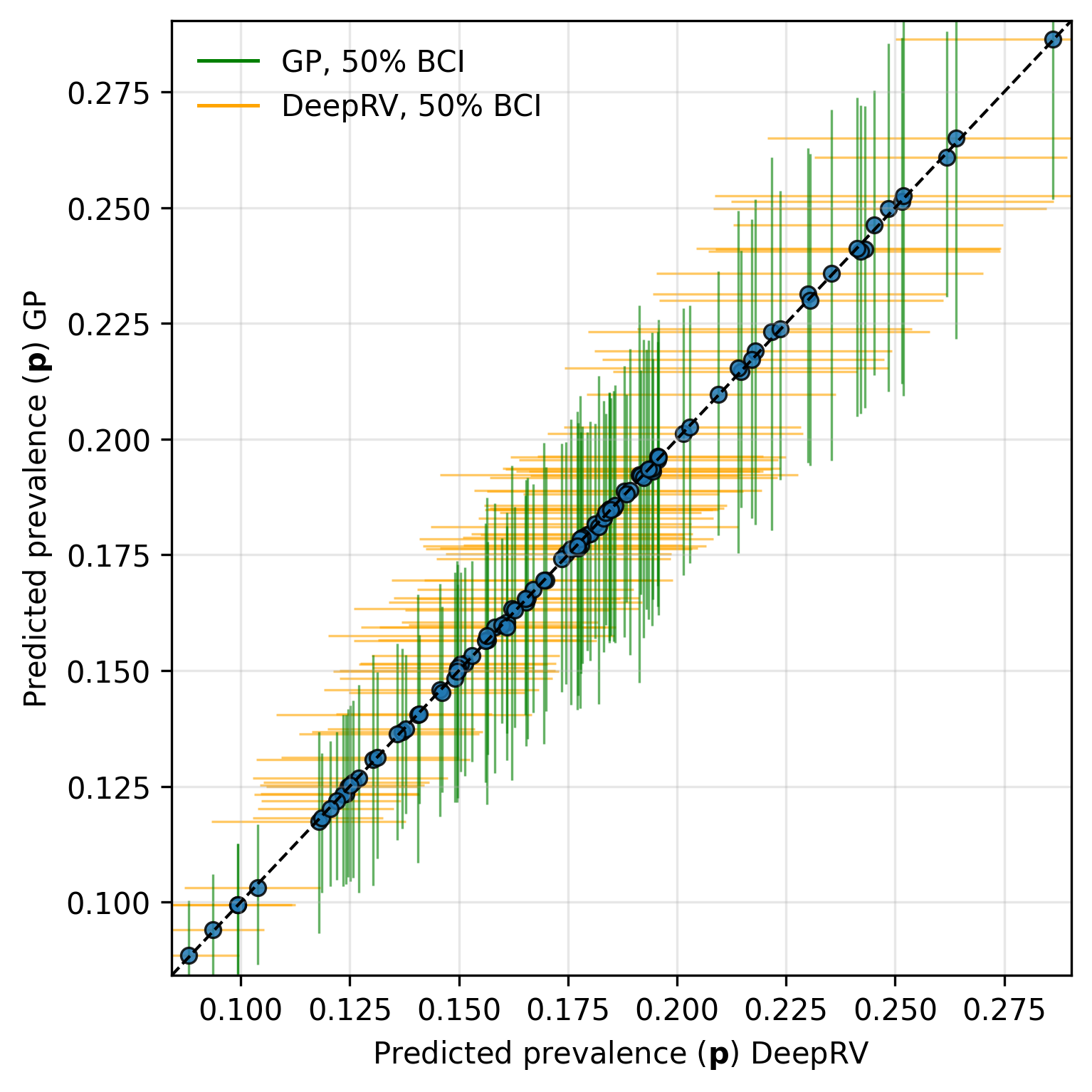}
\caption{Predicted prevalence at 100 randomly selected MSOAs.}
\label{fig:msoa_scatter_m_v_m}
\end{figure}

We ran 4 chains with 4{,}000 warmup and 4{,}000 posterior samples (as in Section~\ref{subsec:benchmarking}). To assess robustness, we randomly masked 50\% of observations. LSOA-level predictive means are shown in \autoref{fig:lsoa_obs_means}. Model-vs-model comparisons of predicted prevalence (DeepRV vs.\ GP) are shown in \autoref{fig:msoa_scatter_m_v_m} (MSOA) and Appendix \autoref{fig:lsoa_scatter_m_v_m} (LSOA). Comparisons against observed prevalence at unobserved MSOA locations are provided in \autoref{fig:msoa_scatter} and \autoref{fig:lsoa_scatter}. We modelled the number of deprived households using a simple binomial likelihood:
\begin{align}\label{equation:Binom_GP_Eq}
\boldsymbol{\theta} :=& \ell, \sigma \sim p_\mathcal{\boldsymbol{\theta}}(\cdot), \notag\\
\mathbf{z} &\sim \mathcal{N}(0, \mathbf{I}),\\
\mathbf{f}_{\boldsymbol{\theta}} =& \operatorname{DeepRV}(\mathbf{z}, \boldsymbol{\theta}),\\
\beta &\sim \mathcal{N}(0, 1),\notag\\
\mathbf{p} =& \operatorname{logit}^{-1}(\beta + \mathbf{f}_{\boldsymbol{\theta}}),\\
\mathbf{y} \sim& \operatorname{Binomial}(\mathbf{N}, \mathbf{p}),\notag
\end{align}
where $\mathbf{N}$ denotes the number of households in each LSOA. Across these checks, DeepRV closely matches the GP in both predicted prevalence and uncertainty on this real-world dataset. A full LSOA GP run would require approximately $\sim$70 hours on our hardware, whereas DeepRV completed in about 3 hours, enabling high-fidelity inference at city scale.

\subsection{Arbitrary Locations DeepRV}\label{subsec: multi location}
We next assess DeepRV’s ability to generalize across datasets with varying numbers and arbitrarily chosen locations. In this setting, both the placement of locations is arbitrary (uniformly sampled) and the number of inputs can change. To emulate GPs in this setting, we use the transformer-based DeepRV, which naturally handles variable-length inputs. As mentioned in Section \ref{subsec:trans}, we employ ID positional embeddings, which in turn requires to specify a priori the maximum number of locations that can be inferred. This design makes it possible to train once and then apply the model to any new set of locations, up to the specified maximum.

We follow the same Matérn-1/2 kernel and Poisson likelihood setup as in \autoref{subsec:benchmarking}, but increase model capacity to four layers, add RFF positional encodings  and train for $2$M steps. Inference was then performed on three datasets of randomly sampled locations in $[0,100]$ with $N=512$, $1024$, and $2048$, benchmarking against both a GP and inducing points.

The results in \autoref{tab:multi_location} show that DeepRV closely matches the GP baseline across predictive and parameter metrics on arbitrary locations. However, to handle this more complex task, the transformer is larger and slower, yielding only modest speed gains ($\approx$10\%). Additional information, including posterior distribution comparisons, and positional encodings ablations that underscores the importance of the ID embeddings for this experiment are provided in Appendix \autoref{subsec:appendix multi_loc}.

\begin{table}[h!]
\scriptsize
\setlength{\tabcolsep}{1.5pt}
\centering
\begin{tabular}{lcccc}
\toprule
Model & MSE($\mathbf{\hat{y}}_{gp}, \mathbf{\hat{y}}$) & Wass($\hat{\ell}_{gp}, \hat{\ell}$) & LPD & Cover-80\% \\
\midrule
GP & - & - & \textbf{-2.00 ± 0.08} & \textbf{0.97 ± 0.01} \\
DeepRV & \textbf{0.01 ± 0.01} & \textbf{0.66 ± 0.20} & \textbf{-2.00 ± 0.08} & \textbf{0.97 ± 0.01} \\
Inducing Pts & 1.82 ± 1.10 & 3.88 ± 0.15 & -2.09 ± 0.10 & 0.86 ± 0.02 \\
\bottomrule
\end{tabular}
\caption{Arbitrary-locations experiment results: (a) Posterior predictive MSE relative to GP; (b) Wasserstein distance between inferred and GP lengthscale posteriors; (c) Log predictive density (LPD); (d) Coverage of the 80\% posterior predictive. Results are averaged across dataset sizes, with the standard error reported.}
\label{tab:multi_location}
\end{table}
\begin{table}[h!]
\centering
\setlength{\tabcolsep}{1.5pt}
\scriptsize
\caption{Architecture ablation for accuracy metrics.}
\label{tab:ablation-acc}
\begin{tabular}{llll}
\toprule
Model & MSE($\mathbf{\hat{y}}_{gp}, \mathbf{\hat{y}}$) & Test Loss & Wass($\hat{\ell}_{gp}, \hat{\ell}$) \\
\midrule
PriorCVAE & 4.424 ± 1.332 & 0.1610 ± 0.0187 & 18.342 ± 4.673 \\
DeepRV-MLP & 0.399 ± 0.143 & 0.0430 ± 0.0017 & 5.545 ± 1.014 \\
DeepRV-gMLP & \textbf{0.005 ± 0.002} & \textbf{0.001 ± 0.000} & \textbf{0.308 ± 0.088} \\
DeepRV-Trans & 0.033 ± 0.012 & 0.004 ± 0.000 & 0.921 ± 0.320 \\
\bottomrule
\end{tabular}
\end{table}
\begin{table}[h!]
\caption{Architecture ablation for efficiency metrics.}
\centering
\setlength{\tabcolsep}{1.5pt}
\scriptsize
\label{tab:ablation-eff}
\begin{tabular}{llll}
\toprule
Model & $\text{ESS} (\ell)/{\text{sec}}$ & Infer Time (s) & Train Time (s) \\
\midrule
GP & 5.035 ± 0.553 & 265.94 ± 26.64 & - \\
PriorCVAE & \textbf{36.174 ± 5.475} & \textbf{60.94 ± 3.19} & 150.6 ± 0.29 \\
DeepRV-MLP & 7.536 ± 1.040 & 73.03 ± 7.25 & \textbf{128.5 ± 0.32} \\
DeepRV-gMLP & 12.709 ± 1.324 & 102.66 ± 9.29 & 283.7 ± 0.42 \\
DeepRV-Trans & 6.474 ± 0.684 & 156.83 ± 12.91 & 2200.5 ± 0.50 \\
\bottomrule
\end{tabular}
\end{table}
\subsection{Ablation Study}\label{subsec: ablation}
We evaluate our architectural choices in DeepRV by comparing DeepRV--MLP, DeepRV--gMLP, DeepRV--Transformer with kernel attention, PriorCVAE, and a GP. We followed the same setup as \autoref{subsec:benchmarking} on a fixed $512$-point 2D grid, across four kernels (Matérn 1/2, 3/2, 5/2, RBF) and three random seeds. 

The results in \autoref{tab:ablation-acc} and \autoref{tab:ablation-eff} show that the performance gap between PriorCVAE and DeepRV--MLP stems from the decoder-only design rather than architectural complexity, as both use the same MLP backbone. 
The transformer variant matches gMLP accuracy but at much higher computational cost, restricting it to the variable-location setting. Overall, all DeepRV variants approximate full GP inference well, with gMLP offering the best trade-off between accuracy and efficiency.

\section{Conclusion}
We presented \textbf{DeepRV}, a decoder-only neural surrogate for Gaussian processes that maps latent draws and kernel parameters directly to function values. Across simulated spatial benchmarks, a non-separable spatiotemporal setting, and a city-scale application (London LSOA), DeepRV consistently matched full GP inference in both predictions and hyperparameter recovery while substantially accelerating MCMC-based inference and retaining modelling flexibility. Compared with popular scalable alternatives (INLA, inducing points, RFFs, VAEs), DeepRV offered the strongest fidelity to exact GPs at practical runtimes on a single GPU, and extended naturally to a transformer variant. The main limitations are the cubic pre-training cost to generate supervision and the assumption of a deterministic mapping from randomness to outputs. Future work will focus on reducing pre-training cost (\textit{e.g.}, Flash or Flex Attention \citep{flash1,flash2,flash3,flex}), improving transfer to unseen resolutions and geometries, and extending the paradigm to broader classes of stochastic processes and probabilistic simulators.

\subsubsection*{Acknowledgments}
J.N. and E.S. acknowledge support in part by the AI2050 program at Schmidt Sciences (Grant [G-22-64476]). J.N. and S.F. acknowledge the EPSRC (EP/V002910/2). D.J. acknowledges his Google DeepMind scholarship. We thank Tom Rainforth for advice on the method and James Bennett for advice on data access. We thank Paolo Andrich and Samir Bhatt for their feedback on the manuscript. 

\bibliography{references}
\clearpage
\onecolumn
\appendix
\section*{Checklist}

\begin{enumerate}

  \item For all models and algorithms presented, check if you include:
  \begin{enumerate}
    \item A clear description of the mathematical setting, assumptions, algorithm, and/or model. [Yes.]
    \item An analysis of the properties and complexity (time, space, sample size) of any algorithm. [Yes.]
    \item (Optional) Anonymized source code, with specification of all dependencies, including external libraries. [Not Applicable. We provide a publicly available code repository.]
  \end{enumerate}

  \item For any theoretical claim, check if you include:
  \begin{enumerate}
    \item Statements of the full set of assumptions of all theoretical results. [Not Applicable.]
    \item Complete proofs of all theoretical results. [Not Applicable.]
    \item Clear explanations of any assumptions. [Not Applicable.]     
  \end{enumerate}

  \item For all figures and tables that present empirical results, check if you include:
  \begin{enumerate}
    \item The code, data, and instructions needed to reproduce the main experimental results (either in the supplemental material or as a URL). [Yes. The paper can be fully reproduced through the provided codebase. Link to the repository can be found in the abstract and in Section \ref{sec:experiments}.]
    \item All the training details (e.g., data splits, hyperparameters, how they were chosen). [Yes. Full training details can be found in the appendix section of each experiment, and in the provided codebase.]
    \item A clear definition of the specific measure or statistics and error bars (e.g., with respect to the random seed after running experiments multiple times). [Yes.]
    \item A description of the computing infrastructure used. (e.g., type of GPUs, internal cluster, or cloud provider). [Yes. This information can be found in the appendix section of each experiment.]
  \end{enumerate}

  \item If you are using existing assets (e.g., code, data, models) or curating/releasing new assets, check if you include:
  \begin{enumerate}
    \item Citations of the creator If your work uses existing assets. [Yes.]
    \item The license information of the assets, if applicable. [Yes. We use only open-source codebases and publicly available data. The licenses of third-party software are specified in their respective repositories, our code is released publicly on GitHub under its repository license, and the ONS data are distributed under their stated public terms.]
    \item New assets either in the supplemental material or as a URL, if applicable. [Yes. The codebase URL is provided in the paper, and the maps used to generate the real-life data experiment \autoref{subsec:london_lsoa} can be downloaded using the instructions within the codebase.]
    \item Information about consent from data providers/curators. [Not Applicable.]
    \item Discussion of sensible content if applicable, e.g., personally identifiable information or offensive content. [Not Applicable.]
  \end{enumerate}

  \item If you used crowdsourcing or conducted research with human subjects, check if you include:
  \begin{enumerate}
    \item The full text of instructions given to participants and screenshots. [Not Applicable.]
    \item Descriptions of potential participant risks, with links to Institutional Review Board (IRB) approvals if applicable. [Not Applicable.]
    \item The estimated hourly wage paid to participants and the total amount spent on participant compensation. [Not Applicable.]
  \end{enumerate}

\end{enumerate}

\clearpage

\section{Benchmark GP Approximations}
\label{app:gp-approximations}

This appendix provides the explicit mathematical formulations of the GP approximations used to benchmark DeepRV.
\subsection{Integrated Nested Laplace Approximation (INLA)}
\label{app:inla}

Integrated Nested Laplace Approximation (INLA) \citep{Rue2009INLA} provides a deterministic alternative to MCMC for latent Gaussian models (LGMs) by approximating posterior marginals.

\paragraph{Latent Gaussian model.}
An LGM with latent field $\mathbf{f}$ and hyperparameters $\boldsymbol{\theta}$ has joint density of:
\begin{equation}
p(\mathbf{y},\mathbf{f},\boldsymbol{\theta})
=
p(\boldsymbol{\theta})\,p(\mathbf{f}\mid\boldsymbol{\theta})
\prod_{i=1}^N p(y_i\mid f_i,\boldsymbol{\theta}),
\qquad
\mathbf{f}\mid\boldsymbol{\theta}\sim\mathcal{N}\!\big(\mathbf{0},\mathbf{Q}(\boldsymbol{\theta})^{-1}\big),
\end{equation}
where $\mathbf{Q}(\boldsymbol{\theta})$ is sparse.

\paragraph{INLA approximation.}
Let $\tilde{p}_G(\mathbf{f}\mid\boldsymbol{\theta},\mathbf{y})$ be a Gaussian approximation to $p(\mathbf{f}\mid\boldsymbol{\theta},\mathbf{y})$ centred at its mode $\mathbf{f}^*(\boldsymbol{\theta})$. INLA approximates the hyperparameter posterior \citep{Rue2009INLA} by:
\begin{equation}
\tilde{p}(\boldsymbol{\theta}\mid\mathbf{y})
\propto
\left.
\frac{p(\mathbf{y},\mathbf{f},\boldsymbol{\theta})}
{\tilde{p}_G(\mathbf{f}\mid\boldsymbol{\theta},\mathbf{y})}
\right|_{\mathbf{f}=\mathbf{f}^*(\boldsymbol{\theta})},
\label{eq:inla_theta}
\end{equation}
and then computes latent marginals by numerical integration over the low-dimensional $\boldsymbol{\theta}$ \citep{Rue2009INLA}.

\paragraph{SPDE--GMRF construction.}
The stochastic partial differential equation (SPDE) approach represents a Mat\'ern Gaussian field as the stationary solution $f(\mathbf{s})$ to a linear stochastic PDE, enabling a sparse Gaussian Markov random field (GMRF) discretisation on a mesh \citep{Lindgren2011SPDE}:
\begin{equation}
(\kappa^2-\Delta)^{\alpha/2}\big(\tau f(\mathbf{s})\big)=\mathcal{W}(\mathbf{s}),
\qquad
\nu=\alpha-\frac{d}{2}.
\end{equation}
Discretising on a triangular mesh gives
\begin{equation}
f(\mathbf{s})\approx\sum_{g=1}^G \psi_g(\mathbf{s})w_g,
\end{equation}
where $\{\psi_g\}$ are local basis functions and $\mathbf{w}=(w_1,\dots,w_G)$ is a Gaussian vector with sparse precision $\mathbf{Q}(\boldsymbol{\theta})$ induced by local mesh neighbourhoods. For our $d=2$ grids we set $\alpha=1.5$, hence $\nu=0.5$, meaning $f(\mathbf{s})$ is approximating the Mat\'ern-$1/2$ kernel.

\paragraph{Code implementation.}
We fit a Poisson log-link model
\begin{equation}
y_i \mid \eta_i \sim \mathrm{Poisson}(\lambda_i),\quad \lambda_i=\exp(\eta_i),\quad
\eta_i=\beta_0+f(\mathbf{s}_i),
\end{equation}
with $\beta_0\sim\mathcal{N}(0,1000)$. We parameterise $\boldsymbol{\theta}=(\theta_1,\theta_2)=(\log\tau,\log\kappa)$ and report the inverse-scale lengthscale
\begin{equation}
\ell=\kappa^{-1}=\exp(-\theta_2).
\end{equation}
We set $\theta_2\sim\mathcal{N}(\mu_2,\sigma_2^2)$ with $(\mu_2,\sigma_2)=(-3.0,0.8)$ and impose an approximate unit-variance constraint by concentrating
\begin{equation}
\theta_1+\theta_2 \sim \mathcal{N}\!\Big(-\tfrac12\log(4\pi),\,\epsilon^2\Big),
\qquad \epsilon=0.05,
\end{equation}
implemented as a bivariate Gaussian prior on $(\theta_1,\theta_2)$ in \texttt{R-INLA}. This prior is constructed so that, under the SPDE parameterisation used in our runs, the implied Mat\'ern marginal variance is approximately fixed to $\sigma^2\approx 1$ (up to the tolerance $\epsilon$), matching the $\sigma^2=1$ convention used in our simulated data. We use \texttt{strategy="laplace"} to form $\tilde{p}_G(\mathbf{f}\mid\boldsymbol{\theta},\mathbf{y})$ via a Laplace approximation around $\mathbf{f}^*(\boldsymbol{\theta})$, and \texttt{int.strategy="grid"} to integrate over $\boldsymbol{\theta}$ on a deterministic grid when computing marginals.

\subsection{Inducing Point Approximation}
\label{app:inducing-points}

We benchmark DeepRV against a classical subset-of-regressors (SoR), also known as the deterministic training conditional (DTC), inducing point approximation \citep{Quinonero2005Unifying,Snelson2006Sparse}.

Let $\mathbf{x}\in \mathbb{R}^N$ denote observation locations and $\mathbf{u}\in \mathbb{R}^M$ inducing locations with $M \ll N$. Define inducing variables
\begin{equation}
\mathbf{f}_u \sim \mathcal{N}(\mathbf{0}, \mathbf{K}_{uu}),\notag
\end{equation}
where $\mathbf{K}_{uu} = k_\theta(\mathbf{u},\mathbf{u})$. The latent GP values at $\mathbf{x}$ are approximated as
\begin{equation}
\mathbf{f} \approx \mathbf{K}_{xu}\mathbf{K}_{uu}^{-1}\mathbf{f}_u,\notag
\end{equation}
with $\mathbf{K}_{xu} = k_\theta(\mathbf{x},\mathbf{u})$.

\paragraph{Code implementation.}
Within our MCMC model, we sample
\[
\mathbf{z} \sim \mathcal{N}(\mathbf{0},\mathbf{I}),
\]
and compute
\begin{align} \mathbf{L}_{uu} = \text{Cholesky}(\mathbf{K}_{uu}), \quad
\bar{\mathbf{f}}_u = \mathbf{L}_{uu}^{-\top}\mathbf{z}, \quad
\mathbf{f} = \mathbf{K}_{xu}\bar{\mathbf{f}}_u .\notag
\end{align}
This is algebraically equivalent to $\mathbf{K}_{xu}\mathbf{K}_{uu}^{-1}\mathbf{z}$, but instead of the explicit matrix inversion we utilize a triangular solver for $\bar{\mathbf{f}}_u$ which was numerically more stable. A small diagonal jitter is added to $\mathbf{K}_{uu}$ prior to Cholesky decomposition. The resulting $\mathbf{f}$ is then used directly inside the likelihood during MCMC.
\subsection{Random Fourier Feature Implementation}
\label{app:rff}

This subsection details the random Fourier feature (RFF) implementations used in our benchmarks. RFFs approximate a stationary kernel $k_\theta(x-x')$ via a finite dimensional feature map $\phi(x)\in\mathbb{R}^M$ such that
\begin{equation}
k_\theta(x,x') \approx \phi(x)^\top \phi(x').
\end{equation}
Following standard constructions \citep{rahimi2007random}, we use cosine features
\begin{equation}
\phi_m(x) = \sqrt{\frac{2}{M}} \cos(\omega_m^\top x + b_m),
\end{equation}
with phases $b_m \sim \mathrm{Uniform}(0,2\pi)$ and frequencies $\omega_m$ drawn from the spectral density of the target kernel.

For each experiments (\textit{e.g.}, for each grid size and lengthscale), the base frequencies $\{\omega_m\}_{m=1}^M$ and phases $\{b_m\}_{m=1}^M$ are sampled once and reused throughout inference. The kernel lengthscale $\ell$ is treated as a latent variable and applied by rescaling $\omega_m \mapsto \omega_m / \ell$.

\paragraph{Kernel-specific sampling.}
We use the following standard spectral constructions \citep{williams2006gaussian}:
\begin{itemize}
\item \textbf{RBF:} $\omega_m \sim \mathcal{N}(\mathbf{0},\mathbf{I})$.
\item \textbf{Matérn-1/2:}
\(
\omega_m = \tan(\pi u_m), \quad u_m \sim \mathrm{Uniform}(-\tfrac{1}{2},\tfrac{1}{2}).
\)
\item \textbf{Matérn-3/2:} Frequencies are sampled using the scale-mixture representation of the Matérn-3/2 spectral density. Let $d$ denote the input dimension. Define
\begin{align}
\nu = \tfrac{3}{2},\quad
\alpha = \frac{2\nu}{\ell^2} = \frac{3}{\ell^2}, \quad
\beta = \nu + \frac{d}{2} = \frac{d+3}{2}.\notag
\end{align}
We sample
\begin{align}
t_m &\sim \mathrm{Gamma}(\beta, \alpha),\quad
\mathbf{z}_m \sim \mathcal{N}(\mathbf{0}, \mathbf{I}_d),\quad
\omega_m = \frac{\mathbf{z}_m}{\sqrt{2 t_m}},\notag
\end{align}
yielding $\omega_m \in \mathbb{R}^d$. Base frequencies are sampled with $\ell=1$ and rescaled by $\ell^{-1}$ during inference.

\end{itemize}

During inference, given $\phi(x)$, the latent function is represented as
\begin{equation}
f(x) = \phi(x)^\top \mathbf{z}, 
\quad \mathbf{z} \sim \mathcal{N}(\mathbf{0},\mathbf{I}),
\end{equation}
yielding a rank-$M$ approximation to the GP prior.
\subsection{Variational Inference Baseline (ADVI)}
\label{app:advi}

We use Automatic Differentiation Variational Inference (ADVI) as implemented in NumPyro\cite{phan2019composable}, with a full-rank Gaussian variational family via the \texttt{AutoMultivariateNormal} guide. ADVI is applied to the same probabilistic model used for GP and DeepRV inference, differing only in the posterior approximation.

Let $\mathbf{z}$ denote the collection of all latent variables and hyperparameters in the model. ADVI approximates the posterior \(
p(\mathbf{z}\mid \mathbf{y})
\) with a multivariate Gaussian variational distribution
\(
q_\lambda(\mathbf{z}) = \mathcal{N}(\boldsymbol{\mu}, \boldsymbol{\Sigma}),
\)
where $\lambda = (\boldsymbol{\mu}, \boldsymbol{\Sigma})$ are variational parameters. These are learned by maximizing the evidence lower bound (ELBO),
\[
\mathcal{L}(\lambda)
=
\mathbb{E}_{q_\lambda(\mathbf{z})}
\big[
\log p(\mathbf{y}, \mathbf{z}) - \log q_\lambda(\mathbf{z})
\big],
\]
using reparameterization gradients.

Optimization is performed for a fixed budget of $50{,}000$ iterations using the Adam optimizer with learning rate $10^{-4}$. After optimization, posterior samples are drawn directly from $q_\lambda(\mathbf{z})$ and used to generate posterior predictive samples through the original model likelihood. 

\subsection{PriorCVAE workflow}

\begin{algorithm*}[h!]
\caption{PriorCVAE \citep{semenova2023priorcvae} workflow}
\begin{algorithmic}
    \STATE Fix the \textbf{spatial structure} of interest $\mathbf{s} = (s_1, \dots, s_n)$, \textit{e.g.} centroids of administrative units
    \STATE Fix the \textbf{latent dimension size} $d\leq n$ for the decoder $D_\psi: \mathbb{R}^d\times\mathcal{C}\rightarrow\mathbb{R}^n$, and the encoder $E_\gamma: \mathbb{R}^n\times\mathcal{C}\rightarrow\mathbb{R}^d$.\vspace{5pt}
    \STATE \textbf{Train PriorCVAE prior}:
        \INDSTATE - Sample hyperparameters: $\boldsymbol{\theta} \sim p_\mathcal{\boldsymbol{\theta}}(\cdot).$
        \INDSTATE - Sample GP realizations: $\mathbf{f}_{\boldsymbol{\theta}} \sim \mathcal{GP}_{\boldsymbol{\theta}}(\cdot)$, over the spatial structure $\mathbf{s}$
        \INDSTATE - Encode $\hat{\mathbf{z}}_\mu,\hat{\mathbf{z}}_\sigma = E_\gamma(\mathbf{f}_{\boldsymbol{\theta}},{\boldsymbol{\theta}})$, sample $\hat{\mathbf{z}} \sim \mathcal{N\textit{}}(\hat{\mathbf{z}}_\mu,\hat{\mathbf{z}}_\sigma)$, and decode $\hat{\mathbf{f}}_{\boldsymbol{\theta}} = D_\psi(\hat{\mathbf{z}},{\boldsymbol{\theta}})$. 
        \INDSTATE - Back propagate the loss:$\mathcal{L}_\text{CVAE} = \frac{1}{\sigma_\text{vae}^2}\operatorname{MSE}(\mathbf{f}_{{\boldsymbol{\theta}}}, \mathbf{\hat{f}}_{\boldsymbol{\theta}}) + \text{KL}\left[\mathcal{N}(\hat{\mathbf{z}}_\mu,\hat{\mathbf{z}}_\sigma) || \mathcal{N}(\mathbf{0}, \mathbf{1})\right]$\vspace{5pt}

    \STATE \textbf{Perform Bayesian inference with MCMC} of the overarching model, including latent variables and hyperparameters ${\boldsymbol{\theta}}$, by approximating $f_{\boldsymbol{\theta}}$ with $\hat{\mathbf{f}}_{{\boldsymbol{\theta}}}$ in a drop-in manner using the trained decoder:\[\mathbf{f}_{\boldsymbol{\theta}} \approx \mathbf{\hat{f}}_{{\boldsymbol{\theta}}} = D_\psi(\mathbf{z}, {\boldsymbol{\theta}}), \mathbf{z} \sim \mathcal{N}(\mathbf{0}, \mathbf{I}_d)\]
\end{algorithmic}
\label{alg:PriorCVAE}
\end{algorithm*}
\clearpage
\section{Experiments}\label{sec:Experiments appendix}

\subsection{Benchmarking DeepRV}\label{subsec:appendix Benchmarking DeepRV}

\subsubsection{Experimental details}

\textbf{Models and architectures.} 
DeepRV variants included a two-layer MLP with ReLU activations, a two-layer gMLP without bottleneck, and a transformer with kernel-based attention bias. PriorCVAE used a standard MLP encoder–decoder. Inducing points and RFFs were implemented in NumPyro. INLA was run with the \texttt{R-INLA} package.

\textbf{Training setup.} 
DeepRV and PriorCVAE were trained with batch size 32 using Optax optimizers with cosine-annealed learning rates and gradient clipping ($\|\cdot\|_2 \leq 3$). DeepRV–gMLP used AdamW with maximum learning rate $10^{-3}$ ($N \leq 32^2$) or $2 \times 10^{-3}$ for larger grids. DeepRV–MLP and PriorCVAE used learning rates of $10^{-3}$ for small grids and up to $5 \times 10^{-3}$ otherwise. Training ran for 200{,}000 steps (300{,}000 steps for $48^2$ and $64^2$ grids). ADVI optimization was performed with Adam at a fixed learning rate of $10^{-4}$ for 50{,}000 steps.

\textbf{Priors.} 
For the Matérn-1/2 kernel, the lengthscale prior was $\ell \sim \text{LogNormal}(3.0, 0.4)$, the variance was fixed at 1, and $\beta \sim \mathcal{N}(0,1000)$. For the Matérn-3/2 kernel, the prior was $\ell \sim \text{LogScaleTransform}(\text{Beta}(4,1))$ spanning $(1,100)$, the variance was fixed at 1, and $\beta \sim \mathcal{N}(0,1)$. 

\textbf{Hardware.} 
All Matérn-1/2 experiments were run on a single NVIDIA GeForce RTX 5090 GPU. Matérn-3/2 experiments used an NVIDIA RTX 5000 Ada GPU. INLA computations were performed on a Mac CPU.

\textbf{Training times.} 
Average training times (in seconds) for DeepRV and PriorCVAE across grid sizes are reported in Table~\ref{tab:train_times_benchmark}. Each entry shows the mean ± standard error over three runs.

\begin{table}[h!]
\centering
\begin{tabular}{llllllllll}
\toprule
Metric & Grid & DeepRV-MLP & DeepRV-gMLP & PriorCVAE \\
\midrule
 & 256 & 163.33 ± 1.24 & 247.29 ± 1.60 & 183.26 ± 1.51 \\
 & 576  & 165.71 ± 1.43 & 249.85 ± 1.44 & 185.69 ± 2.57 \\
Train Time (s) & 1024 & 165.86 ± 1.20 & 360.01 ± 0.17 & 185.97 ± 1.29 \\
 & 2304 & 448.37 ± 0.23 & 1297.65 ± 0.32 & 632.58 ± 0.39 \\
 & 4096 & 1895.87 ± 2.78 & 4106.42 ± 12.66 & 2999.56 ± 2.94 \\
\bottomrule
\end{tabular}
\caption{Training times (in seconds) for DeepRV and PriorCVAE across grid sizes for the Mat\'ern-1/2 kernel, averaged over three runs with standard error.}
\label{tab:train_times_benchmark}
\end{table}

\subsubsection{Results: Mat\'ern-3/2}

\begin{figure*}[t]
    \centering
    \small
    \begin{subfigure}[b]{0.49\textwidth}
        \includegraphics[width=\textwidth]{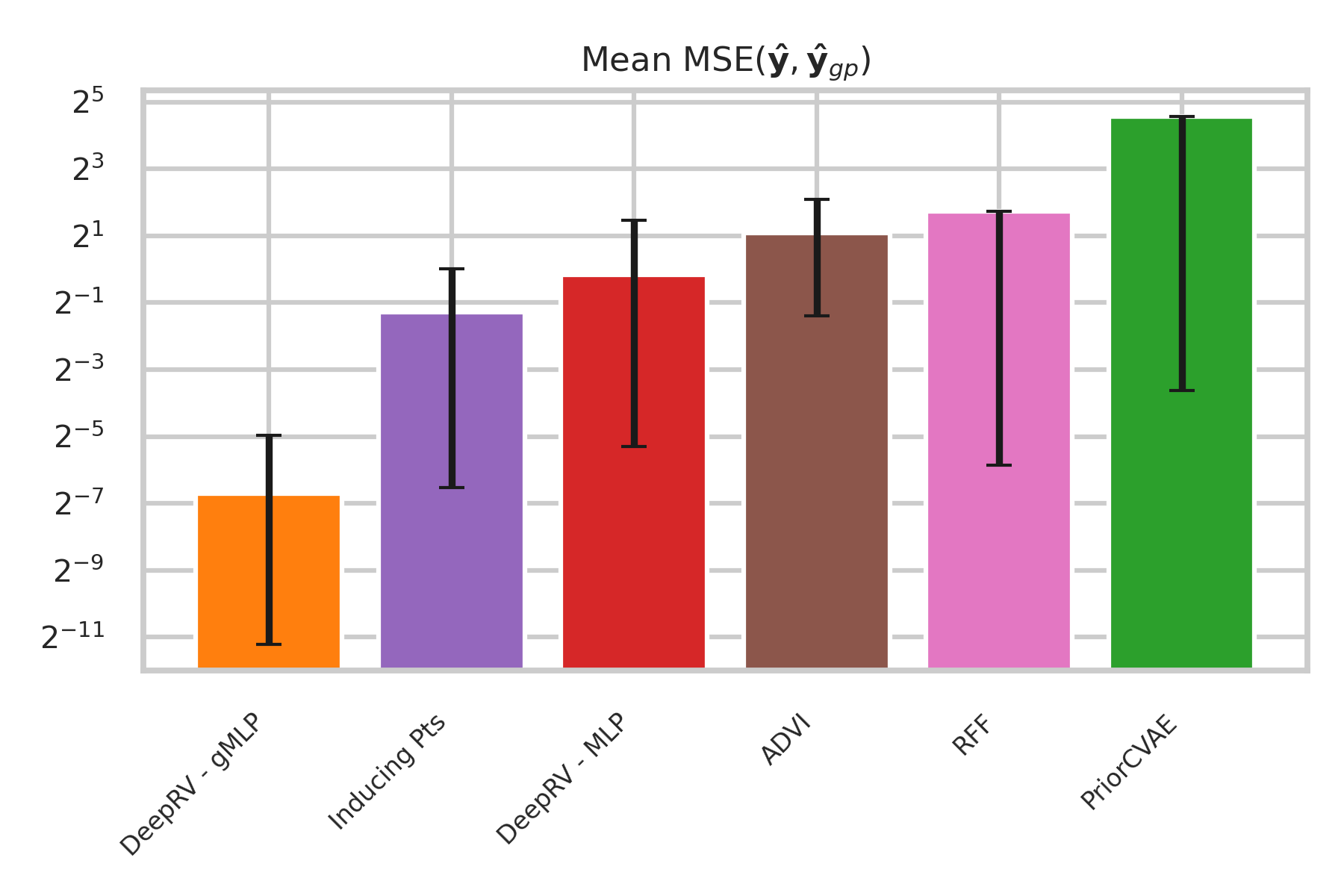}
    \end{subfigure}
    \begin{subfigure}[b]{0.49\textwidth}
        \includegraphics[width=\textwidth]{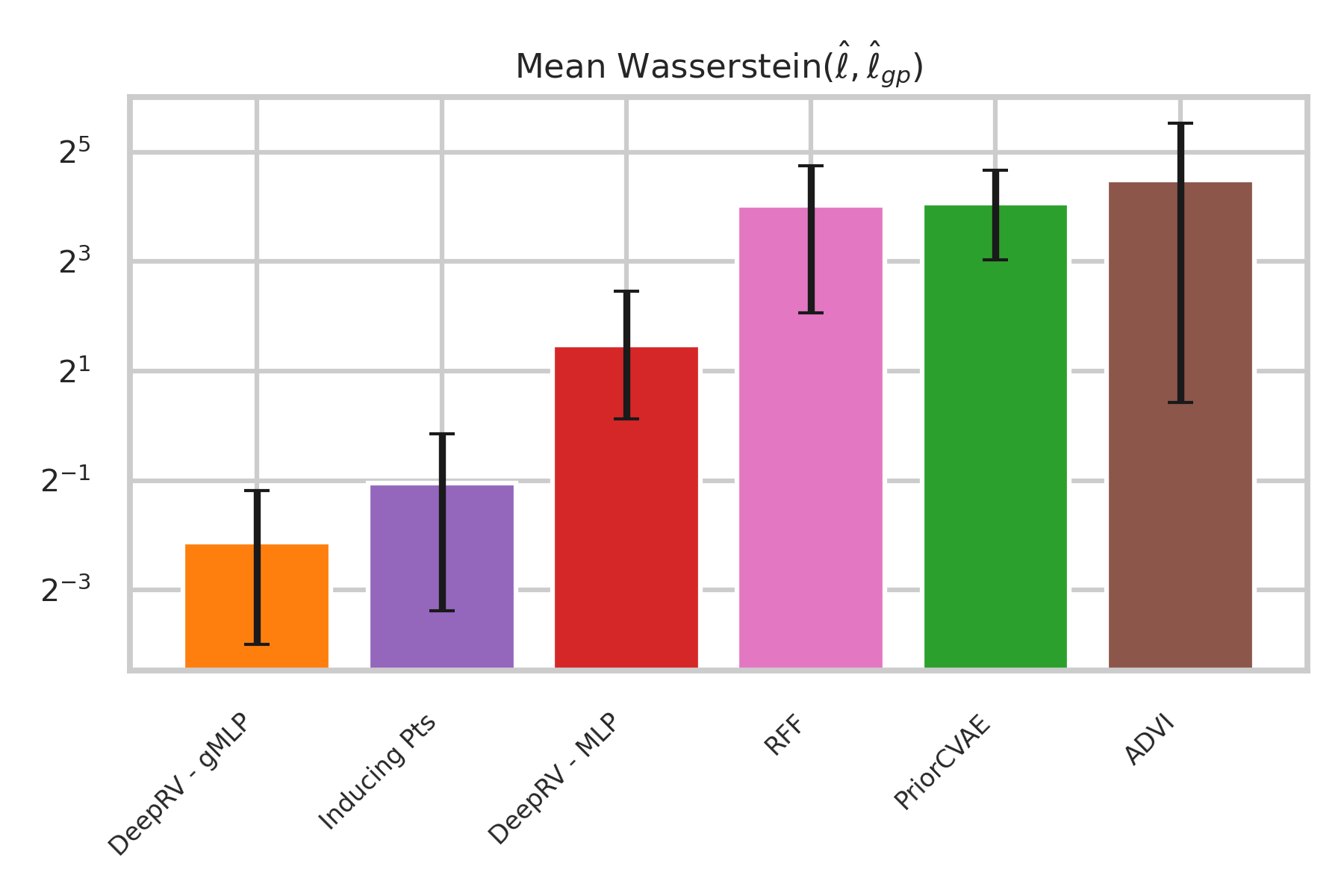}
    \end{subfigure}
    \caption{(a) Posterior predictive MSE relative to full GP MCMC; (b) Wasserstein distance between inferred and full GP MCMC lengthscale posteriors. Y axis is $\operatorname{log}_2$-scaled, to allow a clear comparison of all methods. Results are averaged across true lengthscales and grid sizes over 15 runs, with 10\% and 90\% quantiles reported.}
    \label{fig:bar_summary_main_matern_3_2}
\end{figure*}

We repeated the benchmarking experiment using the Matérn-3/2 kernel, which is not directly supported by standard \texttt{R-INLA}. Details of the setup follow Section~\ref{subsec:benchmarking}, with the modified prior described above. Observations were generated from the Poisson model in Eq.~\ref{equation:Poissoin_GP_Eq}, and inference was performed with NUTS \citep{hoffman2014no} (4 chains for $32^2$, one chain otherwise; 4{,}000 warmup steps and 10{,}000 posterior samples). True lengthscales $\ell \in \{10,30,50\}$ and $\beta=1$ were used, with $\sim$50\% of observations masked in spatially contiguous regions. Results presented in \autoref{tab:res_matern_3_2},\autoref{fig:bar_summary_main_matern_3_2} and  are consistent with the Mat\'ern-1/2 results. Here Inducing Points are able to approximate the GP  better as the kernel is smoother.

\begin{table}[t]
\scriptsize
\centering
\setlength{\tabcolsep}{1.5pt}
\begin{tabular}{ll|cccccccc}
\toprule
metric & Grid & ADVI & GP & DeepRV-MLP & DeepRV-gMLP & Inducing Pts & PriorCVAE & RFF \\
\midrule
 & 256 & 0.667 ± 0.38 & - & 1.216 ± 1.17 & 0.002 ± 0.00 & 0.380 ± 0.35 & 106.077 ± 105.60 & 0.223 ± 0.13 \\
 & 576 & 1.183 ± 0.55 & - & 1.213 ± 0.91 & 0.006 ± 0.00 & 0.397 ± 0.24 & 3.383 ± 1.63 & 1.655 ± 1.03 \\
MSE($\mathbf{\hat{y}}_{gp}, \mathbf{\hat{y}}$) & 1024 & 1.463 ± 0.61 & - & 0.853 ± 0.76 & 0.014 ± 0.01 & 0.804 ± 0.78 & 5.704 ± 5.41 & 12.939 ± 12.84 \\
 & 2304 & 1.479 ± 0.37 & - & 0.165 ± 0.13 & 0.002 ± 0.00 & 0.152 ± 0.14 & 1.445 ± 1.30 & 0.958 ± 0.91 \\
 & 4096 & 5.755 ± 1.76 & - & 1.000 ± 0.35 & 0.024 ± 0.01 & 0.334 ± 0.29 & 1.395 ± 1.02 & 0.740 ± 0.36 \\
\midrule 
 & 256 & 26.94 ± 15.51 & - & 2.92 ± 1.08 & 0.31 ± 0.14 & 0.66 ± 0.40 & 24.07 ± 2.63 & 14.22 ± 7.70 \\
 & 576 & 15.19 ± 8.15 & - & 4.04 ± 1.67 & 0.20 ± 0.09 & 0.75 ± 0.15 & 13.66 ± 5.22 & 16.74 ± 3.71 \\
Wass($\hat{\ell}_{gp}, \hat{\ell}$) & 1024 & 19.40 ± 11.12 & - & 3.93 ± 1.11 & 0.23 ± 0.13 & 0.49 ± 0.22 & 14.09 ± 5.80 & 28.95 ± 13.22 \\
 & 2304 & 26.44 ± 12.27 & - & 1.79 ± 0.47 & 0.19 ± 0.08 & 0.32 ± 0.12 & 13.37 ± 3.51 & 14.47 ± 3.74 \\
 & 4096 & 24.25 ± 11.27 & - & 1.29 ± 0.14 & 0.22 ± 0.07 & 0.19 ± 0.05 & 18.92 ± 5.62 & 6.97 ± 2.82 \\
\midrule 
 & 256 & 326.11 ± 5.45 & 12.66 ± 6.27 & 17.31 ± 9.25 & 18.95 ± 8.49 & 20.68 ± 12.62 & 37.98 ± 20.08 & 2.04 ± 1.06 \\
 & 576 & 199.35 ± 1.04 & 5.54 ± 2.22 & 9.91 ± 5.58 & 14.05 ± 5.29 & 12.25 ± 4.30 & 18.19 ± 13.45 & 15.66 ± 14.95 \\
$\text{ESS } (\ell)/{\text{sec}}$ & 1024 & 483.24 ± 19.58 & 2.40 ± 0.67 & 10.24 ± 1.87 & 10.12 ± 2.22 & 8.27 ± 2.24 & 21.45 ± 11.64 & 0.44 ± 0.27 \\
 & 2304 & 33.06 ± 0.63 & 0.33 ± 0.11 & 1.94 ± 0.44 & 4.79 ± 1.11 & 2.98 ± 0.62 & 4.94 ± 2.41 & 2.13 ± 1.92 \\
 & 4096 & 8.79 ± 0.04 & 0.05 ± 0.03 & 0.65 ± 0.18 & 1.24 ± 0.53 & 2.38 ± 0.97 & 1.29 ± 0.66 & 0.13 ± 0.12 \\
\midrule 
 & 256 & 8 ± 0.13 & 364 ± 109.52 & 138 ± 48.00 & 188 ± 43.79 & 433 ± 269.17 & 100 ± 15.04 & 219 ± 16.85 \\
 & 576 & 12 ± 0.08 & 1294 ± 466.29 & 230 ± 45.33 & 381 ± 87.95 & 391 ± 117.26 & 213 ± 32.01 & 304 ± 9.72 \\
Infer Time (s) & 1024 & 21 ± 0.80 & 2686 ± 781.70 & 350 ± 60.23 & 583 ± 131.98 & 456 ± 111.84 & 304 ± 69.96 & 497 ± 63.00 \\
 & 2304 & 75 ± 1.46 & 14197 ± 2781.03 & 1176 ± 106.86 & 984 ± 178.77 & 892 ± 119.61 & 946 ± 27.88 & 1606 ± 4.14 \\
 & 4096 & 285 ± 0.38 & 139912 ± 64389.16 & 3950 ± 969.34 & 5923 ± 3301.25 & 1792 ± 745.92 & 2920 ± 657.59 & 7273 ± 2625.99 \\
\bottomrule
\end{tabular}
\caption{Mat\'ern-3/2 benchmarking results: (a) Posterior predictive MSE relative to full GP MCMC; (b) Wasserstein distance between inferred and full GP MCMC lengthscale posteriors; (c) Effective $\ell$ sample size (ESS) per second; (d) Inference time in seconds. Results are shown for each grid size and are averaged across the three true lengthscales (10, 30, 50) over 15 runs, with the standard error reported.}\label{tab:res_matern_3_2}
\end{table}

\subsection{DeepRV flexibility: non-separable spatiotemporal kernel}\label{subsec:appendix Non-separable spatiotemporal}

\subsubsection{Experimental details}

\textbf{Models and architectures.} 
We used a two-layer gMLP variant of DeepRV. PriorCVAE employed a standard MLP encoder–decoder. Inducing points, ADVI, and baseline GP were also implemented for comparison.

\textbf{Training setup.} 
DeepRV and PriorCVAE were trained with batch size 32 for 500{,}000 steps. Training used the same optimizers as in the benchmarking experiment: AdamW (cosine-annealed learning rate, gradient clipping $\|\cdot\|_2 \leq 3$) for DeepRV, and Yogi for PriorCVAE. ADVI optimization was performed with Adam at a fixed learning rate of $10^{-4}$ for 50{,}000 steps.  

\textbf{Priors.} 
For inference we used:
\[
\ell \sim \text{LogScaleTransform}(\text{Beta}(4,1)), \,\, 
a \sim \text{LogNormal}(0,1), \,\, 
\alpha \sim \text{Beta}(2,2), \,\,
\nu \sim \text{Uniform}(D,2D), \,\, 
\beta \sim \mathcal{N}(0,1),
\] 
with variance fixed at 1.0. Data were generated with hyperparameters $\ell=20.0$, $\beta=1.0$, $a=0.5$, $\alpha=0.8$, $b=1.0$, $\nu=1.0$, and a dimensionality of $D=2$.

\textbf{Hardware.} 
Experiments were run on a single NVIDIA GeForce RTX 5090 GPU, consistent with the Matérn-1/2 benchmarks.  

\textbf{Training times.} 
Training times (in seconds) are shown in Table~\ref{tab:train_times_spatiotemp}. Each entry is the mean ± standard error across three runs.

\begin{table}[h!]
\centering
\begin{tabular}{lcc}
\toprule
Model & Train time (s) & Infer time (s) \\
\midrule
Baseline GP & – & 5751.35 \\
Inducing Points & – & 965.91 \\
PriorCVAE & 837.38 & 986.12 \\
ADVI & – & 105.48 \\
DeepRV-gMLP & 1164.62 & 1099.13 \\
\bottomrule
\end{tabular}
\caption{Training and inference times (in seconds) for the non-separable spatiotemporal kernel. Train times are reported where models required pre-training. Inference times are reported for all models.}
\label{tab:train_times_spatiotemp}
\end{table}
\subsubsection{Results}

\begin{figure}[h!]
    \centering
    \small
    \includegraphics[width=\textwidth]{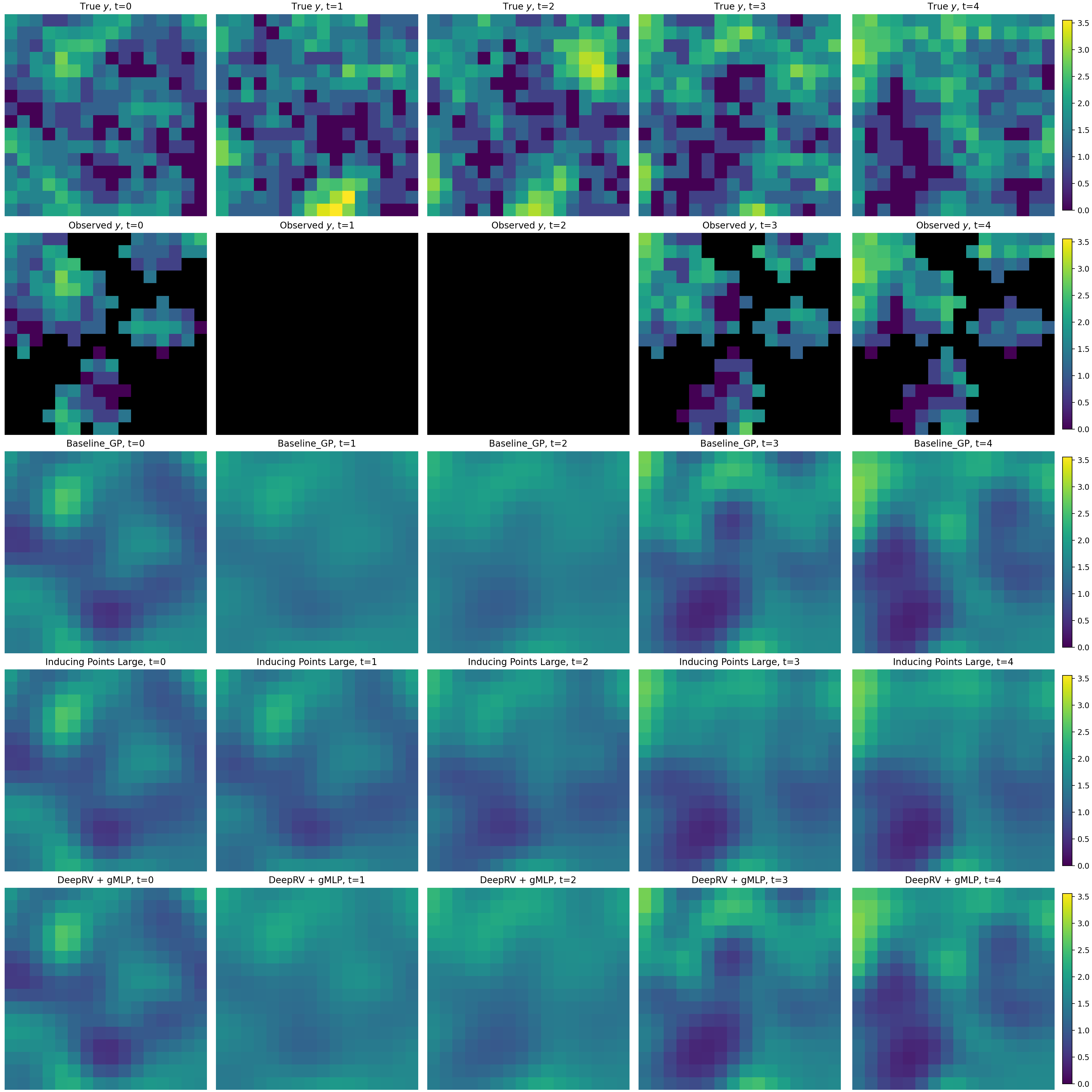}
    \caption{Non-separable spatiotemporal kernel posterior predictives. Results for the top models are presented across time steps.}
\label{fig:obs_means_spatio_temporal}
\end{figure}

The resulting posterior predictive of the top models are shown in Figure~\ref{fig:obs_means_spatio_temporal}. DeepRV closely tracks the GP baseline across space and time, while inducing points and PriorCVAE exhibit higher deviations.

\subsection{Real-world application: London LSOA dataset}\label{subsec:appendix london_lsoa}

\subsubsection{Experimental details}

\textbf{Models and architectures.} 
We used a two-layer gMLP variant of DeepRV. No other approximations (\textit{e.g.},\ inducing points, PriorCVAE) were benchmarked in this experiment; comparisons were made only against the GP baseline.

\textbf{Training setup.} 
DeepRV was trained with batch size 16 for 500{,}000 steps using the AdamW optimizer with cosine-annealed learning rate schedule and gradient clipping ($\|\cdot\|_2 \leq 3$).  

\textbf{Priors.} 
For training we used $\ell \sim \text{Uniform}(1.0, 55.0)$, with variance fixed at 1.0. We slightly tightened the usual prior from $\text{Uniform}(1.0, 55.0)$, as the MAP values of lengthscale were relatively small. Thus, this prior supports lengthscale between $0-50$, and is expansive enough for this dataset.
For inference, priors were centered at MAP values from initialization:
\[
\sigma^2 \sim \text{LogNormal}(\log(\max(\text{var}_{\text{MAP}},10^{-3})), 0.75), \quad 
\ell \sim \text{Gamma}(4, 4/\ell_{\text{MAP}}), \quad 
\beta \sim \mathcal{N}(\beta_{\text{MAP}}, 1.0).
\]

\textbf{Hardware.} 
Experiments were run on a single NVIDIA RTX 5000 Ada GPU, consistent with the Matérn-3/2 benchmarks.  

\textbf{Training and inference times.} 
Training required approximately 12{,}885 seconds ($\sim$3.6 hours) at the LSOA level and 1{,}371 seconds ($\sim$23 minutes) at the MSOA level. Inference required 9{,}081 seconds at LSOA and 3{,}009 seconds at MSOA. For validation, the MSOA full GP was run with 4 chains of 4{,}000 warmup and 4{,}000 posterior samples, while the LSOA short GP calibration run used 2 chains with 1{,}000 warmup and 500 posterior samples.

\subsubsection{Results}

\begin{figure}[h!]
\centering
\includegraphics[width=0.35\textwidth]{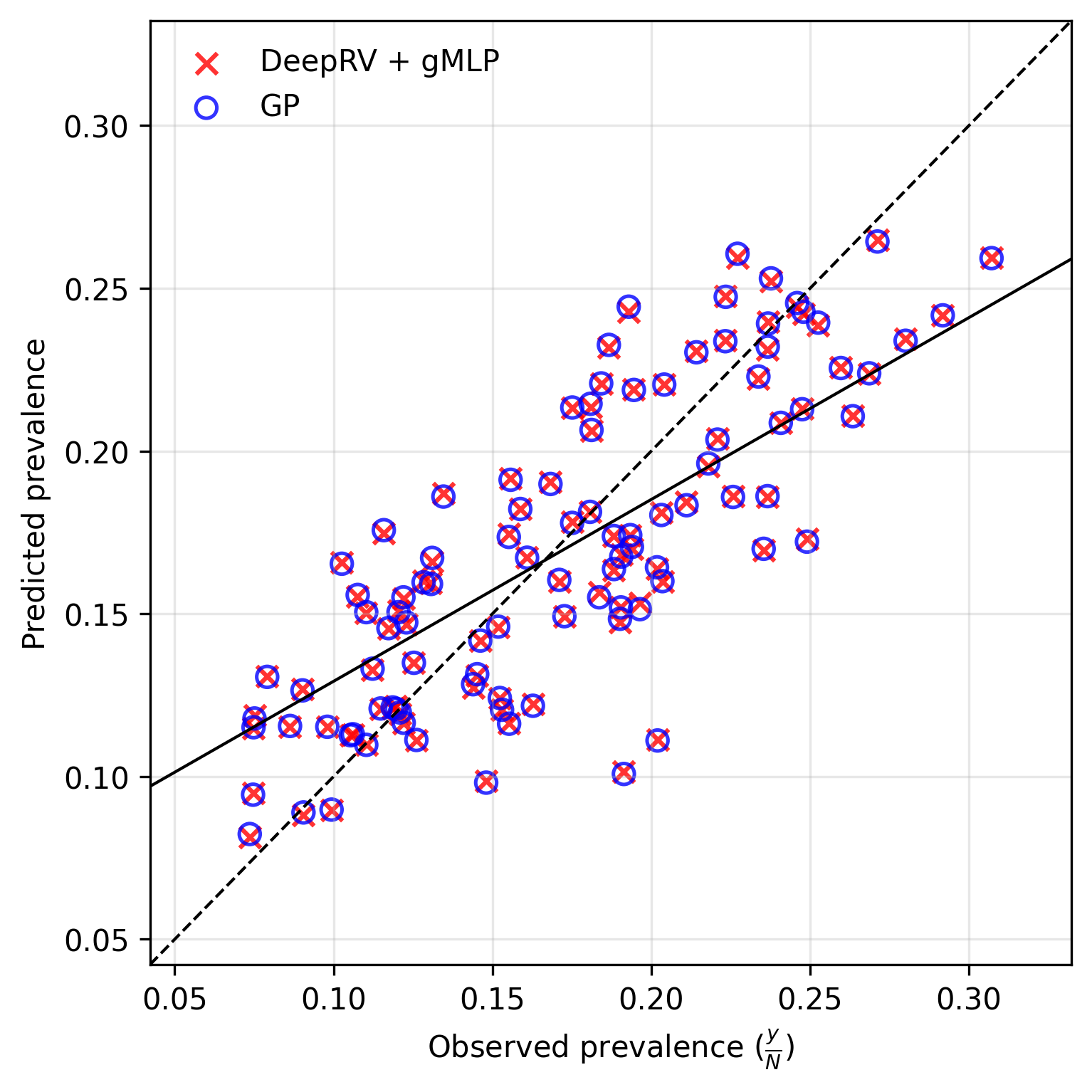}
\caption{Observed versus predicted prevalence ($\mathbf{p}$ in Equation \ref{equation:Binom_GP_Eq}) at 100 randomly selected unobserved MSOA locations. Each point represents one MSOA. The black full line shows the linear regression of DeepRV predictions, illustrating the smoothing effect of the model while maintaining fidelity to the full GP MCMC's prevalence.}
\label{fig:msoa_scatter}
\end{figure}

\begin{figure}[h!]
\centering
\includegraphics[width=0.35\textwidth]{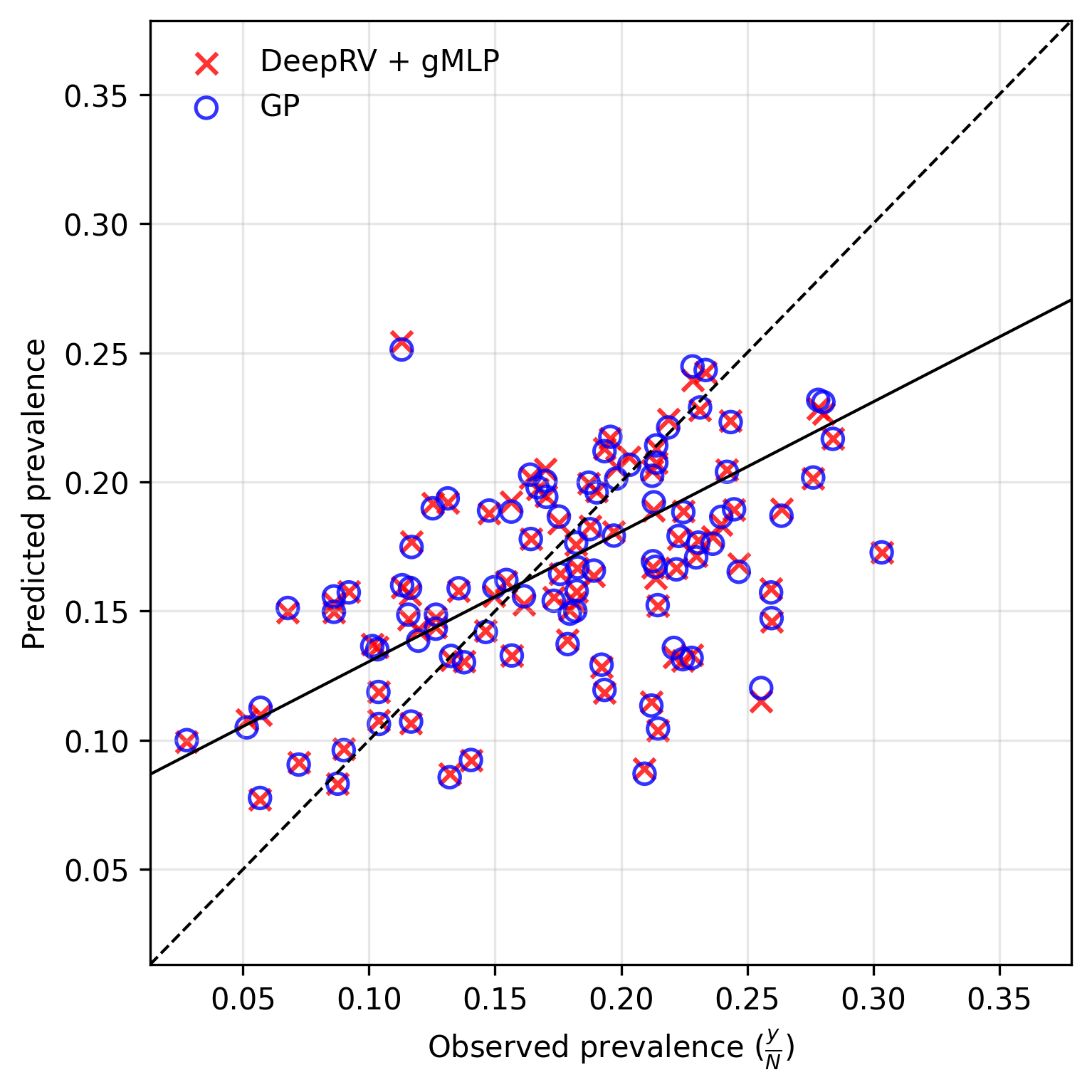}
\caption{Observed versus predicted prevalence ($\mathbf{p}$ in Equation \ref{equation:Binom_GP_Eq}) at 100 randomly selected unobserved LSOA locations. Each point represents one LSOA. The black full line shows the linear regression of DeepRV predictions, illustrating the smoothing effect of the model while maintaining fidelity to the full GP MCMC's prevalence.}
\label{fig:lsoa_scatter}
\end{figure}

\begin{figure}[h!]
\centering
\includegraphics[width=0.4\textwidth]{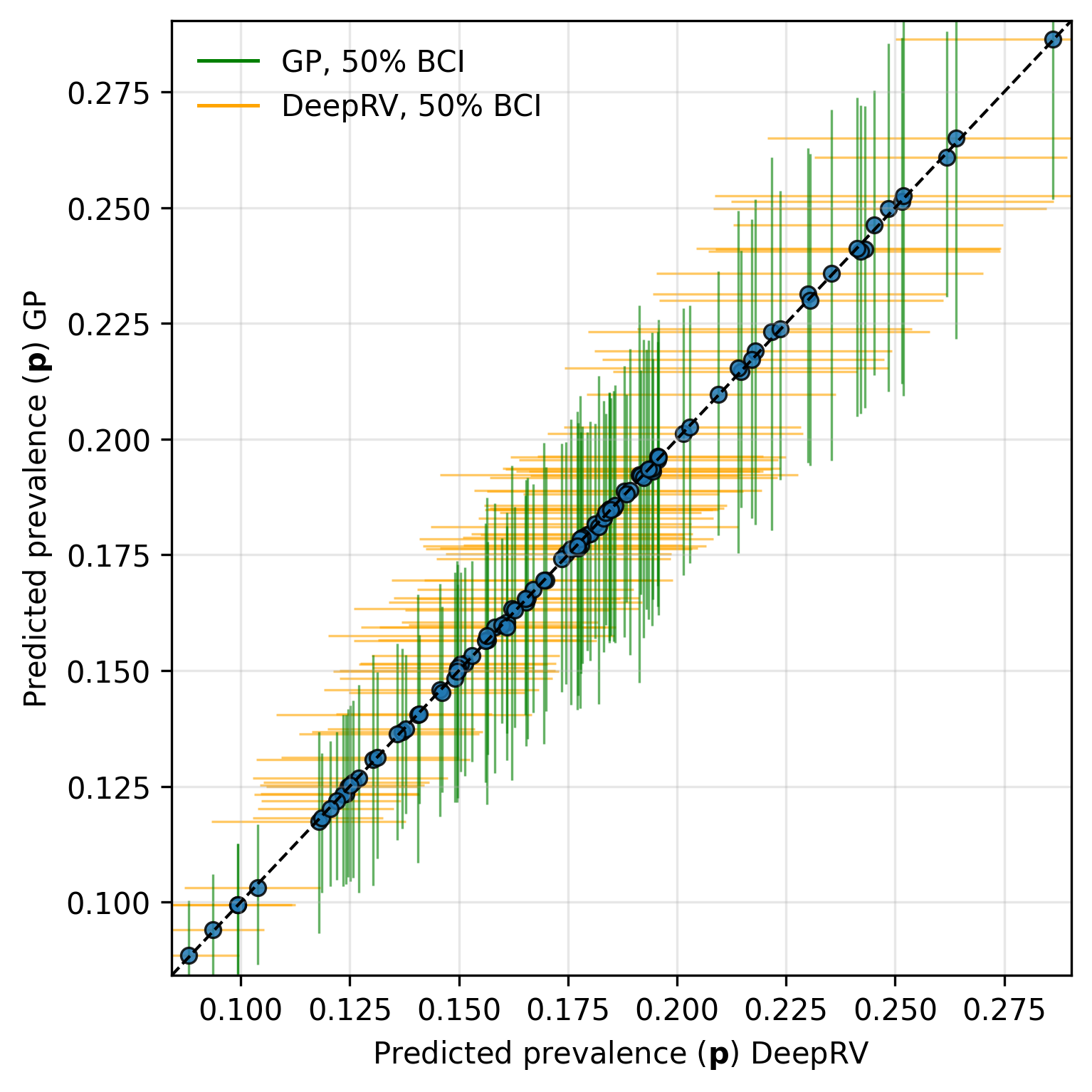}
\caption{Predicted prevalence ($\mathbf{p}$ in Equation \ref{equation:Binom_GP_Eq}) from DeepRV compared against the full GP baseline at 100 randomly selected locations. Each point represents one LSOA. Vertical and horizontal lines denote 50\% credible intervals of models.}
\label{fig:lsoa_scatter_m_v_m}
\end{figure}

Observed versus predicted prevalence comparisons for unobserved locations are shown in Figures~\ref{fig:msoa_scatter} and \ref{fig:lsoa_scatter}. Model-vs-model comparisons of DeepRV against the GP baseline are shown in Figure~\ref{fig:lsoa_scatter_m_v_m}. These results confirm that DeepRV produces predictions and uncertainty estimates closely aligned with the GP baseline at both MSOA and LSOA levels.

\subsection{Arbitrary-Locations DeepRV}\label{subsec:appendix multi_loc}

\subsubsection{Experimental details}

\textbf{Models and architectures.} 
We used a Transformer-based DeepRV with four layers, embedding dimension $D=128$, four attention heads, kernel-attention bias, and identity embeddings. This architecture allows the model to handle arbitrary sets of locations up to a maximum specified at training time. Baselines included the full GP and inducing points.

\textbf{Training setup.} 
DeepRV was trained with batch size 8 for 2M steps using the AdamW optimizer with learning rate $10^{-4}$, cosine annealing, and gradient clipping ($\|\cdot\|_2 \leq 3$). Inducing points were trained with 600k steps. GP required no pre-training. Additionally, to validate the claim that the transformer DeepRV variant could not learn without ID embeddings we provide ablation results for this model with: (a) only RFF positional embeddings without ID embeddings, and (b) only fixed Sinusoidal positional embeddings without ID embeddings. Results for this ablation test are reported in \autoref{tab:multi_location_full}.

\textbf{Priors.} 
For both training and inference, we used $\ell \sim \text{Uniform}(1.0,50.0)$ with variance fixed at 1. The data were generated with true hyperparameters $\ell = 20.0$, $\beta = 1.0$, $\sigma^2 = 1.0$. Inference priors for DeepRV, GP, and Inducing were identical.

\textbf{Hardware.} 
Experiments were run on a single NVIDIA RTX 5090 GPU, consistent with the Matérn-1/2 benchmarks.

\textbf{Training and inference times.} 
DeepRV was trained once, requiring $\sim$43{,}426 seconds ($\approx$12 hours). Inference used two chains with 2{,}000 warmup and 4{,}000 posterior samples. Table~\ref{tab:train_times_multi_loc} reports training and inference times.

\begin{table}[h!]
\centering
\begin{tabular}{lcccc}
\toprule
Model & Train time (s) & Infer time (s) $N=512$ & Infer time (s) $N=1024$ & Infer time (s) $N=2048$ \\
\midrule
GP & – & 644.31 & 2903.27 & 7729.96 \\
Inducing Pts & – & 192.60 & 556.58 & 825.74 \\
DeepRV & 43425.74 & 512.30 & 2680.88 & 7197.59 \\
\bottomrule
\end{tabular}
\caption{Training and inference times (in seconds) for the arbitrary-locations experiment. Inference times are reported per grid size ($N=512, 1024, 2048$). DeepRV was trained once and then applied to all datasets, while GP and Inducing Points require no pre-training.}
\label{tab:train_times_multi_loc}
\end{table}
\subsubsection{Results}

\begin{table}[h!]
\centering
\begin{tabular}{lccccc}
\toprule
Model & Test Loss & MSE($\mathbf{\hat{y}}_{gp}, \mathbf{\hat{y}}$) & Wass($\hat{\ell}_{gp}, \hat{\ell}$) & LPD & Cover-80\% \\
\midrule
GP & - & - & - & \textbf{-2.00 ± 0.08} & \textbf{0.97 ± 0.01} \\
DeepRV & \textbf{0.007} & \textbf{0.01 ± 0.01} & \textbf{0.66 ± 0.20} & \textbf{-2.00 ± 0.08} & \textbf{0.97 ± 0.01} \\
DeepRV RFF only & 0.823 & 0.06 ± 0.04 & 10.07 ± 0.15 & -1.99 ± 0.14 & \textbf{0.98 ± 0.01} \\
DeepRV Sinusoidal only & 0.823 & 0.07 ± 0.05 & 10.16 ± 0.15 & -1.98 ± 0.14 & \textbf{0.98 ± 0.00} \\
Inducing Points & - & 1.82 ± 1.10 & 3.88 ± 0.15 & -2.09 ± 0.10 & 0.86 ± 0.02 \\
\bottomrule
\end{tabular}
\caption{Full Arbitrary-locations experiment results including positional encoding ablations: (a) Posterior predictive MSE relative to GP; (b) Wasserstein distance between inferred and GP lengthscale posteriors; (c) Log predictive density (LPD); (d) Coverage of the 80\% posterior predictive. Results are averaged across dataset sizes, with the standard error reported.}
\label{tab:multi_location_full}
\end{table}

\begin{figure}[t]
    \centering
    \small
    \includegraphics[width=\textwidth]{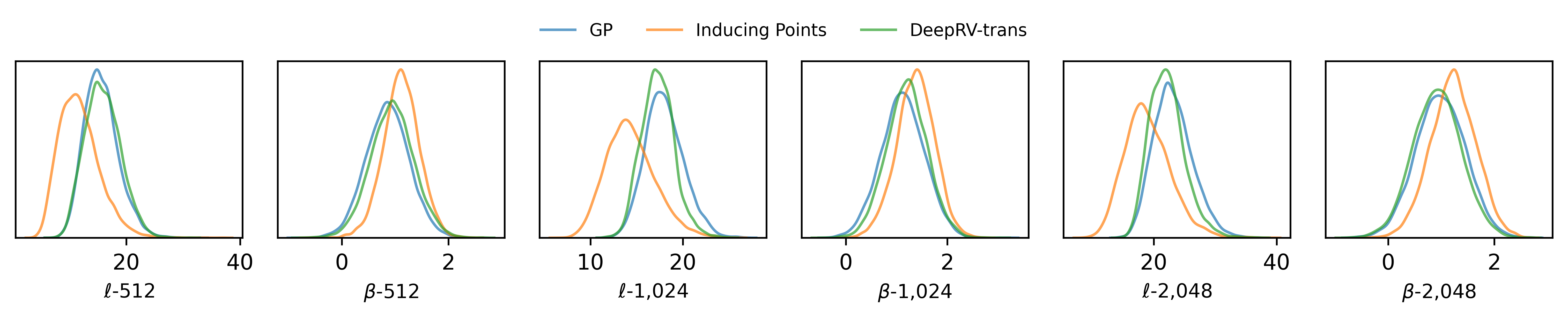}
    \caption{Arbitrary-locations inferred hyperparameter posterior distributions per dataset size ($N=512, 1024, 2048$). DeepRV closely matches GP posteriors across scales, while inducing points deviate.}
\label{fig:appendix_multi_location_posteriors}
\end{figure}

Posterior distribution comparisons across dataset sizes are shown in Figure~\ref{fig:appendix_multi_location_posteriors}. DeepRV closely matches GP posteriors, while inducing points show larger deviations. The ablation results in \autoref{tab:multi_location_full} show that models without ID embeddings fail to learn any structure, collapsing to near-zero function realisations. This produces an artefact of low predictive MSE, despite incorrect inference, as the MCMC overfits the observed data by adjusting $\beta$ and latent $\mathbf{z}$, yielding posterior predictive means similar to the GP baseline while severely misestimating the lengthscale posterior, and failing to infer the underlaying process.

\subsection{Ablation Study}\label{subsec:appendix ablation}

\subsubsection{Experimental details}

\textbf{Models and architectures.} 
We compared a two-layer DeepRV–MLP with ReLU activations, a two-layer DeepRV–gMLP, and a two-layer DeepRV–Transformer with kernel attention and identity embeddings. PriorCVAE used a standard two-layer MLP encoder–decoder. The full GP baseline was also included for comparison.

\textbf{Training setup.} 
All models were trained with batch size 32 for 200{,}000 steps. Optimizers followed the benchmarking setup: AdamW with cosine-annealed learning rate schedule and gradient clipping ($\|\cdot\|_2 \leq 3$) for all DeepRV models except DeepRV–MLP, which used Adam; PriorCVAE used the Yogi optimizer.  

\textbf{Priors.} 
For both training and inference, the lengthscale prior was uniform across the grid $(0,100)$ and $\beta \sim \mathcal{N}(0,1)$. Data were generated from four kernels (Matérn–1/2, Matérn–3/2, Matérn–5/2, and RBF) with hyperparameters drawn from
\[
\ell \sim \text{Uniform}(5,50), \quad \beta \sim \text{Uniform}(0.6,2.0),
\] 
with three seeds per kernel.

\textbf{Hardware.} 
All experiments were run on a single NVIDIA RTX 5000 Ada GPU, consistent with the Matérn–3/2 benchmarks.

\textbf{Training and inference times.} 
Training and inference times were averaged across kernels and seeds. Reported values are provided in Table~\ref{tab:ablation-eff} in the main text.

\subsubsection{Results}

Full ablation results, averaged across kernels and seeds, are reported in the main text (Table~\ref{tab:ablation-acc}). In summary, the gap between PriorCVAE and DeepRV–MLP stems from the decoder-only design, while the Transformer achieves accuracy close to gMLP at substantially higher computational cost. All DeepRV variants closely track the GP baseline in both predictive and parameter inference, with gMLP offering the best balance of accuracy and efficiency. Additionally, we measured the cross-chain accuracy of DeepRV to validate that it remains accurate even when not starting from the same seed as the GP. The results were consistent with the main Table~\ref{tab:ablation-acc}, with the cross chain $\operatorname{MSE}(\hat{\mathbf{y}}, \mathbf{y_{\text{gp}}})$ of $0.012\pm 0.016$, and $\text{Wasserstein}(\hat{{\ell}}, {\ell_{\text{gp}}})$ of $0.272\pm 0.243$.

\subsection{Vecchia Comparison}\label{subsec:appendix_vecchia}

\subsubsection{Experimental details}

\textbf{Models and architectures.} 
We compare three samplers: (i) the full GP process, (ii) a Vecchia approximation, and (iii) a learned gMLP-based DeepRV surrogate.

\textbf{Spatial designs and kernels.} 
We examine the approximations on four kernels, RBF, Matérn-1/2, Matérn-3/2, and Matérn-5/2, for each kernel we generate five independent spatial locations by sampling \(N=2304\) locations uniformly from \([0,100]^2\), enforcing a minimal separation between points. For each spatial design, we draw 1{,}000 independent samples from the GP prior and from the corresponding Vecchia and DeepRV approximations, by sampling lengthscale uniformaly from $0-100$, and fixing variance to $1$. In total we compare the methods on $20,000$ different samples.

\textbf{Vecchia setup.} 
The Vecchia approximation factorises the joint GP prior as
\[
p(\mathbf{f}) \approx \prod_{i=1}^N p\big(f_i \mid \mathbf{f}_{C(i)}\big),
\]
where each conditioning set \(C(i)\) contains at most \(M\) previously ordered locations. For each \(i\), the conditional distribution is Gaussian with
\[
f_i = \mathbf{a}_i^\top \mathbf{f}_{C(i)} + \sqrt{v_i}\,z_i,\qquad
\mathbf{a}_i = K_{C(i),C(i)}^{-1} K_{C(i),i},\qquad
v_i = K_{ii} - K_{i,C(i)} K_{C(i),C(i)}^{-1} K_{C(i),i}.
\]
We set the conditioning size to \(M = 2\sqrt{N}\), matching the computational complexity of DeepRV. Conditioning sets are selected using nearest neighbours in input space, and no specialised ordering strategies are employed. We further enhanced the sampling speed of this approach by parallelising the computation of $\mathbf{a}$ and $\mathbf{v}$, and using \texttt{jax.lax.scan} to compute $\mathbf{f}$ faster.

\textbf{Training setup (DeepRV).} 
DeepRV is trained once per spatial design and kernel using batch size 32 for 300{,}000 optimisation steps with the AdamW optimizer, cosine-annealed learning rate, and gradient clipping (\(\|\cdot\|_2 \leq 3\)). We note that one training run of the RBF kernel did not converge (validation loss $>$ 0.01), and is repeated with a different seed, the resampling of the seed is done automatically, and is integrated in the code.

\textbf{Hardware.} 
All experiments were run on a single NVIDIA RTX 5090 GPU.

\subsubsection{Results}

\begin{table}[h!]
\centering
\small
\begin{tabular}{lccccc}
\toprule
Kernel &
GP time (s) & Vecchia time (s) & DeepRV time (s) & Vecchia MSE & DeepRV MSE \\
\midrule
Matérn-1/2 &
$0.0011 \pm 0.0002$ & $0.0148 \pm 0.0001$ & $\mathbf{0.0002 \pm 0.0000}$ & $\mathbf{0.0000 \pm 0.0000}$ & $\mathbf{0.0000 \pm 0.0002}$ \\
Matérn-3/2 &
$0.0010 \pm 0.0002$ & $0.0150 \pm 0.0001$ & $\mathbf{0.0002 \pm 0.0000}$ & $\mathbf{0.0002 \pm 0.0001}$ & $\mathbf{0.0002 \pm 0.0006}$ \\
Matérn-5/2 &
$0.0011 \pm 0.0002$ & $0.0150 \pm 0.0001$ & $\mathbf{0.0002 \pm 0.0000}$ & $\mathbf{0.0004 \pm 0.0003}$ & $0.0008 \pm 0.0022$ \\
RBF &
$0.0011 \pm 0.0002$ & $0.0150 \pm 0.0001$ & $\mathbf{0.0002 \pm 0.0000}$ & $\mathbf{0.0007 \pm 0.0003}$ & $0.0016 \pm 0.0118$ \\
\bottomrule
\end{tabular}
\caption{Mean $\pm$ standard deviation of sampling time and absolute MSE with respect to a full GP. Results are averaged over 5 spatial locations and 1{,}000 samples per location set.}
\label{tab:vecchia_overall}
\end{table}

As shown in Table~\ref{tab:vecchia_overall}, Vecchia attains very high accuracy relative to the exact GP, in some cases even outperforming DeepRV on the smoother Matérn–5/2 and RBF kernels. However, despite additional implementation-level optimizations, the average time required to generate a single Vecchia sample exceeds that of the dense GP and is approximately two orders of magnitude slower than DeepRV. This behavior is likely due to the fact that dense GP sampling can be fully parallelized on a GPU, whereas Vecchia relies on an inherently sequential computation over conditioning sets, which becomes a bottleneck at this scale. We therefore decided that these runtime costs would make Vecchia comparisons infeasible for the main paper's benchmarking.

\end{document}